\documentclass[nai, crcready]{iosart2x}

\usepackage{amsmath,amsfonts}
\usepackage{algorithmic}
\usepackage{algorithm}
\usepackage{array}
\usepackage{textcomp}
\usepackage{stfloats}
\usepackage{url}
\usepackage{verbatim}
\usepackage{graphicx}
\usepackage{hyperref}
\usepackage[caption=false,font=footnotesize]{subfig}

\usepackage{svg}
\usepackage{stackengine} 
\newcommand\oast{\stackMath\mathbin{\stackinset{c}{0ex}{c}{0ex}{\ast}{\bigcirc}}}
\usepackage{amsmath}
\usepackage{multirow}
\usepackage{booktabs}
\usepackage{amssymb}
\newcommand{\Reals}{\ensuremath{\mathbb{R}}} 
\usepackage{pifont}
\newcommand{\xmark}{\ding{55}}%
\newcommand{\cmark}{\ding{51}}%

\usepackage{graphicx}
\usepackage{comment}
\usepackage{multirow}
\usepackage{import}
\usepackage{wasysym,ifsym}
\usepackage{stmaryrd}
\usepackage{float}
\usepackage[flushleft, para]{threeparttable}
\usepackage[accsupp]{axessibility}  % Improves PDF readability for those with disabilities.

\usepackage{xspace}
\newcommand{\name}{BCF\xspace}
\newcommand{\codename}{GSBC\xspace}

\usepackage[acronym, style=super]{glossaries}

\newacronym{code}{GSBC}{generalized sparse block code}
\glsdisablehyper
\newcommand{\rebuttal}[1]{{\color{black}#1}}

\pubyear{0000}
\volume{0}
\firstpage{1}
\lastpage{1}

\begin{document}
\begin{frontmatter}

%\pretitle{}
\title{Factorizers for Distributed Sparse Block Codes}
\runtitle{Factorizers for Distributed Sparse Block Codes}
%\subtitle{}

% For one author:
%\author{\inits{N.}\fnms{Name1} \snm{Surname1}\ead[label=e1]{first@somewhere.com}}
%\address{Department first, \orgname{University or Company name},
%Abbreviate US states, \cny{Country}\printead[presep={\\}]{e1}}

% Two or more authors:
\begin{aug}
\author[A,B]{\inits{M.}\fnms{Michael} \snm{Hersche}\ead[label=e1]{her@zurich.ibm.com}}
\author[A,B]{\inits{A.}\fnms{Aleksandar} \snm{Terzić}\ead[label=e2]}{eks@zurich.ibm.com}
\author[A]{\inits{G.}\fnms{Geethan} \snm{Karunaratne}\ead[label=e3]{kar@zurich.ibm.com}}
\author[A]{\inits{J.}\fnms{Jovin} \snm{Langenegger}\ead[label=e1]{mail@jovin.ch}}
\author[B]{\inits{A.}\fnms{Ang\'eline} \snm{Pouget}\ead[label=e1]{apouget@ethz.ch}}
\author[A]{\inits{J.}\fnms{Giovanni} \snm{Cherubini}\ead[label=e1]{giovanni.cherubini@hotmail.com}}
\author[B]{\inits{L.}\fnms{Luca} \snm{Benini}\ead[label=e4]{lbenini@iis.ee.ethz.ch}}
\author[A]{\inits{N.-N.}\fnms{Abu} \snm{Sebastian}\ead[label=e5]{ase@zurich.ibm.com}}
\author[A]{\inits{A.}\fnms{Abbas} \snm{Rahimi}\ead[label=e6]{abr@zurich.ibm.com}%
\thanks{Corresponding author. \printead{e6}.}}
\address[A]{IBM Research -- Zurich, S\"{a}umerstrasse 4, 8803 R\"{u}schlikon, Switzerland.}
\address[B]{Department of Information Technology and Electrical Engineering, ETH Z\"{u}rich, Gloriastrasse 35, 8092 Z\"{u}rich, Switzerland.}
%\printead[presep={\\}]{e2,e3}}
\end{aug}

%\begin{review}{editor}
%\reviewer{\fnms{First} \snm{Editor}\address{\orgname{University or Company name}, \cny{Country}}}
%\reviewer{\fnms{Second} \snm{Editor}\address{\orgname{First University or Company name}, \cny{Country}
%    and \orgname{Second University or Company name}, \cny{Country}}}
%\end{review}
%\begin{review}{solicited}
%\reviewer{\fnms{First} \snm{Solicited reviewer}\address{\orgname{University or Company name}, \cny{Country}}}
%\reviewer{\snm{anonymous reviewer}}
%\end{review}
%\begin{review}{open}
%\reviewer{\fnms{First} \snm{Open Reviewer}\address{\orgname{University or Company name}, \cny{Country}}}
%\end{review}

\begin{abstract}
Distributed sparse block codes (SBCs) exhibit compact representations for encoding and manipulating symbolic data structures using fixed-width vectors.
One major challenge however is to disentangle, or factorize, the distributed representation of data structures into their constituent elements without having to search through all possible combinations. 
This factorization becomes more challenging when SBCs vectors are noisy due to perceptual uncertainty and approximations made by modern neural networks to generate the query SBCs vectors.
To address these challenges, we first propose a fast and highly accurate method for factorizing a more flexible and hence generalized form of SBCs, dubbed GSBCs.
Our iterative factorizer introduces a threshold-based nonlinear activation, conditional random sampling, and an $\ell_\infty$-based similarity metric. 
Its random sampling mechanism, in combination with the search in superposition, allows us to analytically determine the expected number of decoding iterations, which matches the empirical observations up to the \codename's bundling capacity. 
Secondly, the proposed factorizer maintains a high accuracy when queried by noisy product vectors generated using deep convolutional neural networks (CNNs).
This facilitates its application in replacing the large fully connected layer (FCL) in CNNs, whereby $C$ trainable class vectors, or attribute combinations, can be implicitly represented by our factorizer having $F$-factor codebooks, each with $\sqrt[\leftroot{-2}\uproot{2}F]{C}$ fixed codevectors.
We provide a methodology to flexibly integrate our factorizer in the classification layer of CNNs with a novel loss function.
With this integration, the convolutional layers can generate a noisy product vector that our factorizer can still decode, whereby the decoded factors can have different interpretations based on downstream tasks.
We demonstrate the feasibility of our method on four deep CNN architectures over CIFAR-100, ImageNet-1K, and RAVEN datasets.  
In all use cases, the number of parameters and operations are notably reduced compared to the FCL.

\end{abstract}


\end{frontmatter}

%%%%%%%%%%% The article body starts:

\section{Introduction}

Vector-symbolic architectures (VSAs)~\cite{MAP_1998, GaylerJackendoff2003, PlateHolographic1995, PlateHolographic2003,KanervaHyperdimensional2009} are a class of computational models that provide a formal framework for encoding, manipulating, and binding symbolic information using fixed-size distributed representations. 
VSAs feature compositionality and transparency, which enabled them to perform analogical mapping and retrieval~\cite{LargePatternsGreatSymbols,AnalRetrieval_2000,AnalMapp_2009}, inductive reasoning~\cite{RasmussenInductiveReasoning2011,EmruliAnalogical2013}, and probabilistic abductive reasoning~\cite{hersche2022nvsa,hersche2023nvsa_nesy}. 
Moreover, the VSA's distributed representations can mediate between rule-based symbolic reasoning and connectionist models that include neural networks.
Recent work~\cite{hersche2022nvsa} has shown how VSA, as a common language between neural networks and symbolic AI, can overcome the binding problem in neural networks and the exhaustive search problem in symbolic AI.

In a VSA, all representations---from atoms to composites---are high-dimensional distributed vectors of the same fixed dimensionality.
% %
An atom in a VSA is a randomly drawn i.i.d. vector that is dissimilar (i.e., quasi-orthogonal) to other random vectors with very high probability, a phenomenon known as concentration of measure~\cite{LedouxConcentration2001}.
Composite structures are created by manipulating and combining atoms with well-defined dimensionality-preserving operations, including multiplicative binding, unbinding, additive bundling (superposition), and permutations.
The binding operation can yield quasi-orthogonal results, which, counterintuitively, can still encode semantic information.
% %
For instance, we can describe a concept in a scene (e.g., a \texttt{black circle}) with two factors (color and shape) by binding quasi-orthogonal atomic vectors ($\mathbf{x}_{\mathrm{black}}\oast \mathbf{x}_{\mathrm{circle}}$). The resulting product vector is quasi-orthogonal to all other possible vectors (atomic and composite). 
Yet, decomposing it into its factors ($\mathbf{x}_{\mathrm{black}}$ and $\mathbf{x}_{\mathrm{circle}}$) reveals the semantic relation between a \texttt{black circle} and a \texttt{black square} since both include $\mathbf{x}_{\mathrm{black}}$.

However, decomposing, or disentangling, a bound product vector into its factors is computationally challenging, requiring checking all the possible combinations of factors.
Extending the previous example from two to $F$ factors, each factor $f$ having a codebook of $M_f$ codevectors, there are $\prod_{f=1}^{F} M_f$ possible combinations to be searched in the product space for factorizing a product vector. 
To alleviate this hard combinatorial search problem, rapid iterative approaches were proposed such as resonator networks~\cite{Resonator1,Resonator2} and follow-up stochastic factorizers with nonlinear activation~\cite{langenegger2022imcfac}.
However, these existing solutions can only infer the factors of dense bipolar product vectors (i.e., each vector element is a Rademacher random variable) and face challenges with other types of VSA representations such as sparse block codes (SBCs)~\cite{RachkovskijBinding2001, laiho2015sparse, FradySDR2021}.
SBCs exhibit compact memory footprint, ideal variable binding properties~\cite{FradySDR2021}, high information capacity for associative memories~\cite{gripon2011sparse, knoblauch2020iterative, schlegel2022comparison}, biological plausibility~\cite{MasseFruitFly2009,HTM2017,WILLSHAW1969, OlshausenSparseAct1996}, and amenability for implementation on emerging neuromorphic hardware~\cite{Bent2022SDRTrueNorth,Renner2022SDRBindingSpike}.
Motivated by these key aspects of SBCs, there is a need to come up with a rapid iterative approach to accurately factorize SBC product vectors.

This paper provides the following contributions, which are divided into two main parts.
In Part~I, for the first time, we propose an iterative block code factorizer (\name) that can reliably factorize blockwise distributed product vectors.
The used codebooks in \name are binary SBCs, which span the product space, while \name can factorize product vectors from a more \gls{code}. 
Hence, factorizing binary SBCs is a special case.
\name introduces a configurable threshold, a conditional sampling mechanism, and a new $\ell_\infty$-based similarity search operation. % for the first time in the VSA context.
During the iterative decoding, the novel sampling mechanism induces a random search in superposition over the product space if no confident solution is present. 
\name improves the convergence speed of the state-of-the-art stochastic factorizer~\cite{langenegger2022imcfac} on a large problem size of $10^6$ by up to 6$\times$. 
To gain a deeper understanding of \name's iterative search in superposition, we leverage its configurable threshold and sampling mechanism to configure it as an unconditional sampler that randomly searches over the product space. 
This allows us to determine the expected number of decoding iterations analytically, which matches the empirical observations when operating within the \gls{code}'s bundling capacity.

In Part~II, we present an application for \name that reduces the number of parameters in fully connected layers (FCLs).
FCLs are ubiquitous in modern deep learning architectures and play a major role by accounting for most of the parameters in various architectures, such as transformers~\cite{ProdKey_NIPS19,FCLareKVmem2021}, extreme classifiers~\cite{XML-CNN2017,ganesan2021learning}, and CNNs for edge devices~\cite{qian2020doweneed}.
\rebuttal{Given an FCL with respective input and output dimensions $D_i$ and $D_o$}, we can replace its trainable parameters $\mathbf{W}\in \Reals^{D_i \times D_o}$ by a \name with $F$ codebooks of fixed parameters, each $\mathbf{X}^f\in \{0,1\}^{D_i \times \sqrt[\leftroot{-2}\uproot{2}F]{D_o}}$.
The structure of the codebooks is naturally given when the product space is defined by semantic attributes (e.g., in RAVEN~\cite{Raven_19}), or can be arbitrarily defined when no semantic attributes are provided (e.g., for natural images in ImageNet-1K~\cite{deng2009imagenet}).
To map sensory inputs to \gls{code} product vectors, we train deep convolutional layers with a novel blockwise additive loss that can directly use \name in place of an FCL classifier.
Our \name reduces the total number of parameters across a wide range of deep CNNs and datasets by 0.5--44.5\% while maintaining a high accuracy within $0$--$4.46\%$ compared to the baseline CNNs using the large FCL.
Our \name also lowers the computational cost of the classifier layer by $55.2$--$86.7\%$ with respect to FCLs.

\section{VSA Preliminary}
VSAs define operations over (pseudo)random vectors with independent and identically distributed (i.i.d.) components.
% codebook
Computing with VSAs begins by defining a basis in the form of a codebook $\mathbf{X}:=\{\mathbf{x}_1,\mathbf{x}_2, ..., \mathbf{x}_M\}:=\{\mathbf{x}_i\}_{i=1}^{M}$.
% Quasi orthogonality 
If the dimension $D_p$ of two randomly drawn vectors of the space is sufficiently large, they are highly likely to have an almost-zero similarity, i.e., they are quasi-orthogonal~\cite{KanervaHyperdimensional2009}. 
% main operations
VSAs use three primary operations---bundling, binding, and permutation---that form an algebra over the space of vectors. 
\rebuttal{The bundling represents a set of vectors via vector addition with a possible nonlinearity, resulting a vector that is similar to all vectors from the set.
In contrast, binding and permutation yield result vectors that are dissimilar to its input vectors. 
}
Combined with a similarity metric, these operations support various cognitive data structures: variable binding, sequence, and hierarchy.
See~\cite{HDCSurvey_PI} for a review.

% Dense representation
For example, consider a VSA model based on the bipolar vector space~\cite{GaylerJackendoff2003}, i.e., $\mathbf{x} \in \left\{-1, +1 \right\}^{D_p}$. 
One can define binding and unbinding in this vector space as the Hadamard (i.e., elementwise) product.
The similarity between two vectors in the space is typically measured using the cosine similarity metric.
A possible bundling operation is the elementwise sum followed by the sign function, setting all negative elements to $-1$ and the positive to $+1$. 
To keep representations bipolar, elements with a sum equal to zero are randomly set to $-1$ or $+1$.

% Binary sparse block codes
As an alternative, binary sparse block codes (binary SBCs)~\cite{laiho2015sparse} induce a local blockwise structure that exhibits ideal variable binding properties~\cite{FradySDR2021} and high information capacity when used in associative memories~\cite{gripon2011sparse, knoblauch2020iterative, schlegel2022comparison}. 
In binary SBCs, the $D_p$-dimensional vectors are divided into $B$ blocks of equal length, $L=D_p/B$, where only one element per block is set to 1. 
The vectors can either be described with a $D_p$-dimensional binary SBC vector (denoted as $\mathbf{x}$), or with a $B$-dimensional offset vector where each element indicates the index of the nonzero element within each block (denoted as $\dot{\mathbf{x}}$).
The vectors are initialized by randomly setting one element in each block to 1.
The bundling of two or more vectors is defined as their elementwise addition, followed by a selection function that retains the sparsity by setting the largest element of each block to 1 and the remaining elements to 0.
The binding of two vectors is the elementwise modulo-$L$ sum of their offset representation.
Similarly, unbinding is defined using the modulo-$L$ difference. 
Both binding and unbinding preserve dimensionality and sparsity. 
A typical choice for the similarity metric is the normalized dot-product, which counts the number of overlapping elements of two vectors~\cite{BinDensity_2018}.

\section{Related Work}
\subsection{Factorizing distributed representations}
The resonator network~\cite{Resonator1, Resonator2} avoids brute-force search through the combinatorial space of possible \rebuttal{factorization} solutions by exploiting the \emph{search in superposition} capability of VSAs.
The iterative search process converges empirically by finding correct factors under operational capacity constraints~\cite{Resonator2}. 
The resonator network can accurately factorize dense bipolar distributed vectors generated by a two-layer perceptron network trained to approximate the multiplicative binding for colored MNIST digits~\cite{Resonator1}.
\rebuttal{Alternatively, the resonator network can also factorize complex-valued product vectors representing a scene encoded via a template-based VSA encoding~\cite{renner2022neuromorphic} or convolutional sparse coding~\cite{kymn_compositional_2024}}. 
However, the resonator network suffers from a relatively low operational capacity \rebuttal{(i.e., the maximum factorizable problem size given a certain vector dimensionality)}, and the limit cycles that impact convergence. 
To overcome these two limitations, a stochastic in-memory factorizer~\cite{langenegger2022imcfac} introduces new nonlinearities and leverages intrinsic noise of computational memristive devices. 
As a result, it increases the operational capacity by at least five orders of magnitude, while also avoiding the limit cycles and reducing the convergence time compared to the resonator network.

Nevertheless, we observed that the accuracy of both the resonator network and the stochastic factorizer notably drops (by as much as 16.22\%) when they are queried with product vectors generated from deep CNNs processing natural images (see Table~\ref{tab:noFC}).  
This challenge motivated us to switch to alternative block code representations instead of dense bipolar, whereby we can retain high accuracy by using our \name.
Moreover, compared to the state-of-the-art stochastic factorizer, \name requires fewer iterations irrespective of the number of factors $F$ (see Table~\ref{tab:dense_vs_sparse}).
Interestingly, it only requires two iterations to converge for problems with a search space as large as $10^4$.

\subsection{Fixing the final FCL in CNNs}
Typically, a learned affine transformation is placed at the end of deep CNNs, yielding a per-class value used for classification. 
In this FCL classifier, the number of parameters is proportional to the number of class categories. 
Therefore, FCLs constitute a large portion of the network’s total parameters: for instance, in models for edge devices, FCLs constitute 44.5\% of ShuffleNetV2~\cite{ShuffleNetV2_ECCV2018}, or 37\% of MobileNetV2~\cite{MobileNetV2_CVPR2018} for ImageNet-1K.
This dominant parameter count is more prominent in lifelong continual learning models, where the number of classes quickly exceeds a thousand and increases over time~\cite{Hersche_2022_CVPR}.

To reduce the training complexity associated with FCLs, various techniques have been proposed to fix their weight matrix during training.
In turn, an FCL is replaced by a Hadamard matrix~\cite{hoffer2018fix}, or a cheaper Identity matrix~\cite{qian2020doweneed}, or vertices of a simplex equiangular tight frame~\cite{NeuralCollaps_NIPS2021}.
Although partly effective, due to square-shaped structures, these methods are restricted to problems in which the number of classes is smaller than or equal to the feature dimensionality, i.e., $D_i=D_o$.
Methods that simply draw class vectors randomly distributed over a hypersphere~\cite{mettes2019hyperspherical,RandomClassVec2021} were proposed to address this limitation.
However, these methods still need to store the individual class vectors, which imposes the FCL's conventional cost of $\mathcal{O}(D_i \cdot D_o)$ for memory storage and compute complexity during training and inference.

Our \name with two factors can reduce the memory and compute complexity to $\mathcal{O}(D_i \cdot \sqrt{D_o})$.
This is done by using randomly-drawn distributed binary SBCs that form an intermediate product vector space whose dimensionality ($D_p$) is notably lower than the number of classes ($D_p\ll D_o$), but high enough such that a large number of classes can be expressed thanks to the supplied quasi-orthogonality.
We show that the product vector space can be built either at the output of the last convolutional layer directly (i.e., its dimensionality is set by the feature dimension $D_p=D_i$), or at the output of a smaller FCL as a projection layer (i.e., its dimensionality can be chosen).
This flexibly enables a trade-off between the number of removable parameters and obtainable accuracy.

\section{Part~I: Factorization of Generalized Sparse Block Codes}

This section presents our first contribution: we propose a novel block code factorizer (\name) that efficiently finds the factors of product vectors based on block codes. 
We first introduce \glsfirst{code}, a generalization of the previously presented binary SBC. 
 We present corresponding binding, unbinding, and bundling operations and a novel similarity metric based on the $\ell_\infty$-distance.
We then continue to the exposition and experimental evaluation of our \name, which is capable of fast and accurate factorization of \gls{code} product vectors.

\subsection{Generalized sparse block codes (GSBCs)}
\label{BCAs}
Like binary SBCs, \glspl{code} divide the $D_p$-dimensional vectors into $B$ blocks of equal length $L=D_p/B$. 
However, the individual blocks are not restricted to be binary or sparse.
The requirements imposed upon the vectors are that their elements are in $\Reals^{+}$, and each block has a unit $\ell_1$-norm.
Binary SBCs satisfy both constraints and are valid GSBCs. 
Fig.~\ref{fig:SBCfactorizer} illustrates an example of a binary SBC and a \gls{code} vector. 
The blockwise distribution of the \gls{code} representation can be interpreted as blockwise probabilistic mass functions, serving as a proxy for the binary SBC vector in this example. 
Neural networks can produce such \gls{code} product vectors.
Besides their benefits in the integration with neural networks, the \gls{code} representations inside \name enable more accurate and faster factorization.

The individual operations for the \glspl{code} are defined as follows:
\paragraph{Binding/Unbinding}
We exploit general binding and unbinding operations in blockwise circular convolution and correlation to support arbitrary block representations.
Specifically, if both operands have blockwise unit $\ell_1$-norm, the result does as well.

\paragraph{Bundling}
The bundling of several vectors is defined as their elementwise sum followed by a normalization operation, ensuring that each result block has unit $\ell_1$-norm.

\paragraph{$\ell_\infty$-based Similarity}
We propose a novel similarity measure based on the $\ell_\infty$-norm of the elementwise difference between two \gls{code} vectors $\mathbf{x}_i$ and $\mathbf{x}_j$: 
\begin{align}
    s_{\infty}(\mathbf{x}_i, \mathbf{x}_j) := 1-\ell_\infty(\mathbf{x}_i- \mathbf{x}_j),
\end{align}
with $\ell_\infty(\mathbf{a}) = \max_i|\mathbf{a}[i]|$, where $\mathbf{a}[i]$ denotes the $i$-th element of a vector $\mathbf{a}$.

For any \gls{code} vectors $\mathbf{a}$ and $\mathbf{b}$, it holds that $0\leq\ell_\infty(\mathbf{a}- \mathbf{b})\leq1$. 
Therefore, our novel similarity metric satisfies $0 \leq s_{\infty}(\mathbf{a}, \mathbf{b}) \leq 1$, whereby equality on the right-hand side holds if, and only if, $\mathbf{a}=\mathbf{b}$.
Table~\ref{tab:table1} compares the operations of the \glspl{code} with respect to the binary SBCs.

\begin{table}[]
\caption{Comparison of operations of binary SBCs and our \glspl{code}. All operations except for the similarity are applied blockwise.}
\label{tab:table1}
% \resizebox{\linewidth}{!}{
\begin{tabular}{lcc}
\toprule
             & \textbf{Binary SBCs}\cite{laiho2015sparse}                 & \textbf{\glspl{code}} (ours)     \\
\cmidrule(r){1-1}\cmidrule(r){2-2}\cmidrule(r){3-3}      
Binding ($\oast$) & \begin{tabular}[c]{@{}c@{}}Modulo-$L$ sum \\ of nonzero indices\end{tabular}            &\begin{tabular}[c]{@{}c@{}} Blockwise \\ circular convolution  \end{tabular}                                              \\
\cmidrule(r){1-1}\cmidrule(r){2-2}\cmidrule(r){3-3} 
Unbinding ($\oslash$)  & \begin{tabular}[c]{@{}c@{}}Modulo-$L$ difference\\  between nonzero indices\end{tabular} & \begin{tabular}[c]{@{}c@{}} Blockwise\\ circular correlation   \end{tabular}                                             \\
\cmidrule(r){1-1}\cmidrule(r){2-2}\cmidrule(r){3-3} 
Bundling ($\oplus$)  & Argmax of sum                                                                          & \begin{tabular}[c]{@{}c@{}}Sum \& normalization\end{tabular} \\
\cmidrule(r){1-1}\cmidrule(r){2-2}\cmidrule(r){3-3} 
Similarity  & Dot-product                                                                         & \begin{tabular}[c]{@{}c@{}}$\ell_\infty$-based sim. \\ or dot-product\end{tabular} \\
\bottomrule
\end{tabular}
% }
\end{table}

\subsection{Factorization problem}
We define the factorization problem for \glspl{code} and our factorization approach for two factors. 
Applying our method to more than two factors is straightforward; corresponding experimental results will be presented in Section~\ref{subsec:fac_results}. 

Given two codebooks\footnote{In our experiments, the codebooks consist of binary SBC codewords, but they could be \gls{code} vectors too.}, $\mathbf{X}^1$:=$\lbrace\mathbf{x}_i^1 \rbrace_{i=1}^{M_1}$ and $\mathbf{X}^2$:=$\lbrace\mathbf{x}_i^2 \rbrace_{i=1}^{M_2}$, and a product vector
%\footnote{The product vector can on the other hand be non-binary and dense, e.g. a noise-corrupted binding of several SBC codevectors or a neural network output.} 
$\mathbf{p}=\mathbf{x}_i^1\oast \mathbf{x}_j^2$ formed by binding two factors from the codebooks, we aim to find the estimate factors $\hat{\mathbf{x}}^1 \in \mathbf{X}^1$ and $\hat{\mathbf{x}}^2 \in \mathbf{X}^2$ that satisfy
\begin{align}
    \mathbf{p} = \hat{\mathbf{x}}^1\oast \hat{\mathbf{x}}^2.
\end{align}
A naive brute-force approach would compare the product vector ($\mathbf{p}$) to all possible combinations spanned by the product space $\mathbf{P}=\mathbf{X}^1\oast \mathbf{X}^2 := \left\{\mathbf{x}_1^1\oast \mathbf{x}_1^2, \mathbf{x}_1^1\oast \mathbf{x}_2^2,...,  \mathbf{x}_{M_1}^1\oast \mathbf{x}_{M_2}^2\right\}$. 
This results in a combinatorial search problem requiring $M_1 \cdot M_2$ similarity computations. \name can notably reduce the computational complexity.

\subsection{Block code factorizer (\name)}

\begin{figure*}
    \centering
    \includegraphics[width=\textwidth]{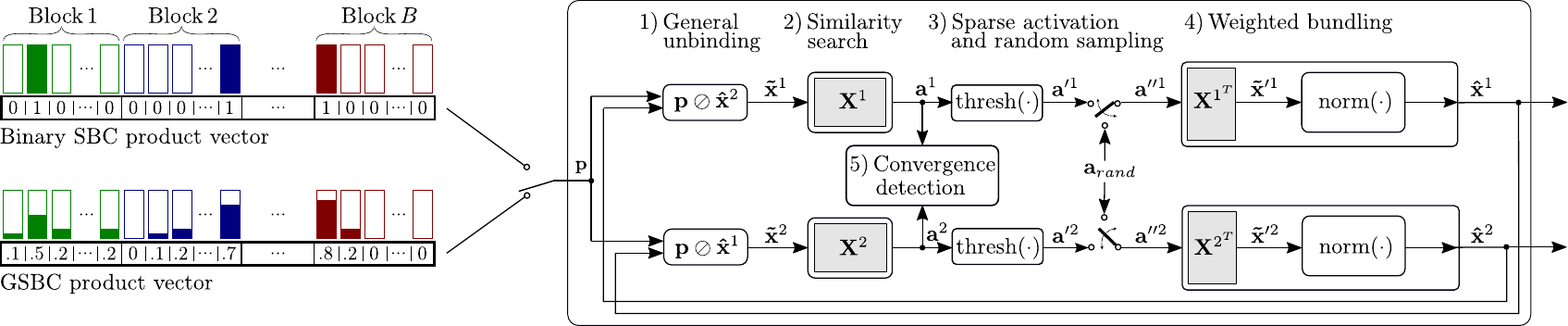}
    \caption{Block code factorizer (\name) for $F=2$ factors. 
    It can factorize both synthetic binary SBC product vectors and \gls{code} product vectors ($\mathbf{p}$) which might result from a neural network mapping.
    }
  \label{fig:SBCfactorizer}
\end{figure*}

Here, we introduce our novel \name that efficiently finds the factors of product vectors based on \glspl{code}, shown in Fig.~\ref{fig:SBCfactorizer}.  
The product vector decoding begins with initializing the estimate factors $\hat{\mathbf{x}}^1(0)$ and $\hat{\mathbf{x}}^2(0)$ by bundling all vectors from the corresponding codebooks. 
Then, the estimate factors are iteratively updated through the following steps. 

\textbf{Step 1: Unbinding.} At  the start of iteration $t\geq1$, the estimate factors from the previous iteration $t-1$ are unbound from the product vector by using blockwise circular correlation:
\begin{align}
    \tilde{\mathbf{x}}^1(t) &= \mathbf{p}\oslash\hat{\mathbf{x}}^2(t-1) \\
    \tilde{\mathbf{x}}^2(t) &= \mathbf{p}\oslash\hat{\mathbf{x}}^1(t-1).
\end{align}

\textbf{Step 2: Similarity search.} Next, we query the associative memory, containing the codebook $\mathbf{X}^f$ for factor $f$ with the unbound factor estimates. 
Here, we deploy our novel $\ell_\infty$-based similarity as an effective associative search.
At iteration $t$, this yields a vector of similarity scores $\mathbf{a}^f(t) \in \Reals^{M_f}$ for each factor $f$. 
The $i$-th element in $\mathbf{a}^f(t)$ is computed as:
\begin{align}
    \mathbf{a}^f(t)[i] = s_\infty(\tilde{\mathbf{x}}^f(t), \mathbf{x}^f_i).
\end{align}
We observe that a conventional dot-product similarity causes a notable performance drop, as shown in Fig.~\ref{fig:factorizer-main-results}.

\textbf{Step 3: Sparse activation and conditional random sampling.} 
Recent work on stochastic factorizers~\cite{langenegger2022imcfac} demonstrated that applying a threshold function to the elements of the similarity vector can improve convergence speed and operational capacity. 
We deploy a similar idea in our \name. 
In this step, the previously computed similarities are compared against a fixed threshold $T \in \Reals^{+}$. 
Similarity values that are larger than the threshold propagate forward, whereas lower ones get zeroed out:
\begin{align}
    \mathbf{a'}^f(t) &= \mathrm{thresh}(\mathbf{a}^f(t);T) \\
    \mathrm{thresh}(\mathbf{a};T)[i] &= \begin{cases}
    \mathbf{a}[i],& \text{if } \mathbf{a}[i]\geq T\\
    0,              & \text{otherwise}.
\end{cases}
\end{align}

This nonlinearity allows us to focus on the most promising solutions by discarding the \rebuttal{presumably} incorrect low-similarity ones. 
However, thresholding entails the possibility of ending up with an all-zero similarity vector, effectively stopping the decoding procedure.
To alleviate this issue, upon encountering an all-zero similarity vector, we randomly generate a subset of equally weighted similarity values:
\begin{align}
    \mathbf{a''}^f(t) = \begin{cases}
    \mathbf{a'}^f(t),& \text{if } \mathbf{a'}^f(t)\neq \mathbf{0}\\
    \mathbf{a}_{rand},              & \text{otherwise,}
\end{cases}
\label{eqn:eq8}
\end{align}
where $\mathbf{a}_{rand} \in \Reals^{M_f}$ is a vector in which $A$-many randomly selected elements are set to $1/A$.
In combination with step 4 (weighted bundling), the conditional random sampling given by Eq.~\eqref{eqn:eq8} yields an equally weighted bundling of $A$ randomly selected codewords, \rebuttal{where $A$ can be interpreted as the sampling width.} 

The novel threshold and conditional sampling mechanisms are simple and interpretable, yet they lead to faster convergence. 
The stochastic factorizer~\cite{langenegger2022imcfac} relied on various noise instantiations at every decoding iteration. 
The necessary stochasticity was supplied from intrinsic noise of phase-change memory devices and analog-to-digital converters of a computational analog memory tile.
Instead, \name remains deterministic in the decoding iterations unless all elements in the similarity vector are zero, in which case it activates only a single random source. 
This can be seen as a conditional \emph{restart} of \name using a new random initialization. 
The conditional random sampling could be implemented with a single random permutation of a seed vector in which $A$-many arbitrary values are set to $1/A$.
The conditional random sampling mechanism is also interpretable, in the sense that it allows to analytically determine the expected number of decoding iterations, subject to the bundling capacity \rebuttal{(i.e., the maximum number of vectors that can be bundled and reliably retrieved). 
Section~\ref{subsec:bcf-ablation} provides empirical insights.}

\textbf{Step 4: Weighted bundling.} Finally, we generate the next factor estimate $\hat{\mathbf{x}}^f(t)$ as the normalized weighted bundling of the factor's codevectors:
\begin{align}
     \hat{\mathbf{x}}^f(t) = \frac{(\mathbf{X}^f)^T \mathbf{a}''^f(t)}{\sum_{i=1}^{M_f} \mathbf{a}''^f(t)[i]}.
\end{align}
The codevectors $\lbrace\mathbf{x}_i^1 \rbrace_{i=1}^{M_1}$ are \glspl{code} with unit $\ell_1$-norm blocks; hence, dividing the weighted bundling by the sum of the weights yields valid \glspl{code}.

\textbf{Step 5: Convergence detection.} The iterative decoding is repeated until \name converges or a predefined maximum number of iterations ($N$) is reached.
We define the maximum number of iterations such that \name does at most as many similarity searches as the brute-force approach~\cite{langenegger2022imcfac}: 
\begin{align}\label{eq:maxiter}
    N := \frac{\prod_{f=1}^{F} M_f}{\sum_{f=1}^F M_f}.
\end{align}
The convergence detection mechanism is based on an additional, fixed threshold. Decoding is stopped as soon as both similarity vectors ($\mathbf{a}^1(t)$ and $\mathbf{a}^2(t)$) contain an element that exceeds a predefined detection threshold value ($T_c$)\cite{langenegger2022imcfac}.
We set it to $T_c=0.8$ for synthetic product vectors and $T_c=0.5$ for noisy product vectors from deep neural networks. 

\begin{figure}[t]
\centering
\subfloat[Optimal threshold ($T^*$).]{\includegraphics[width=0.35\linewidth]{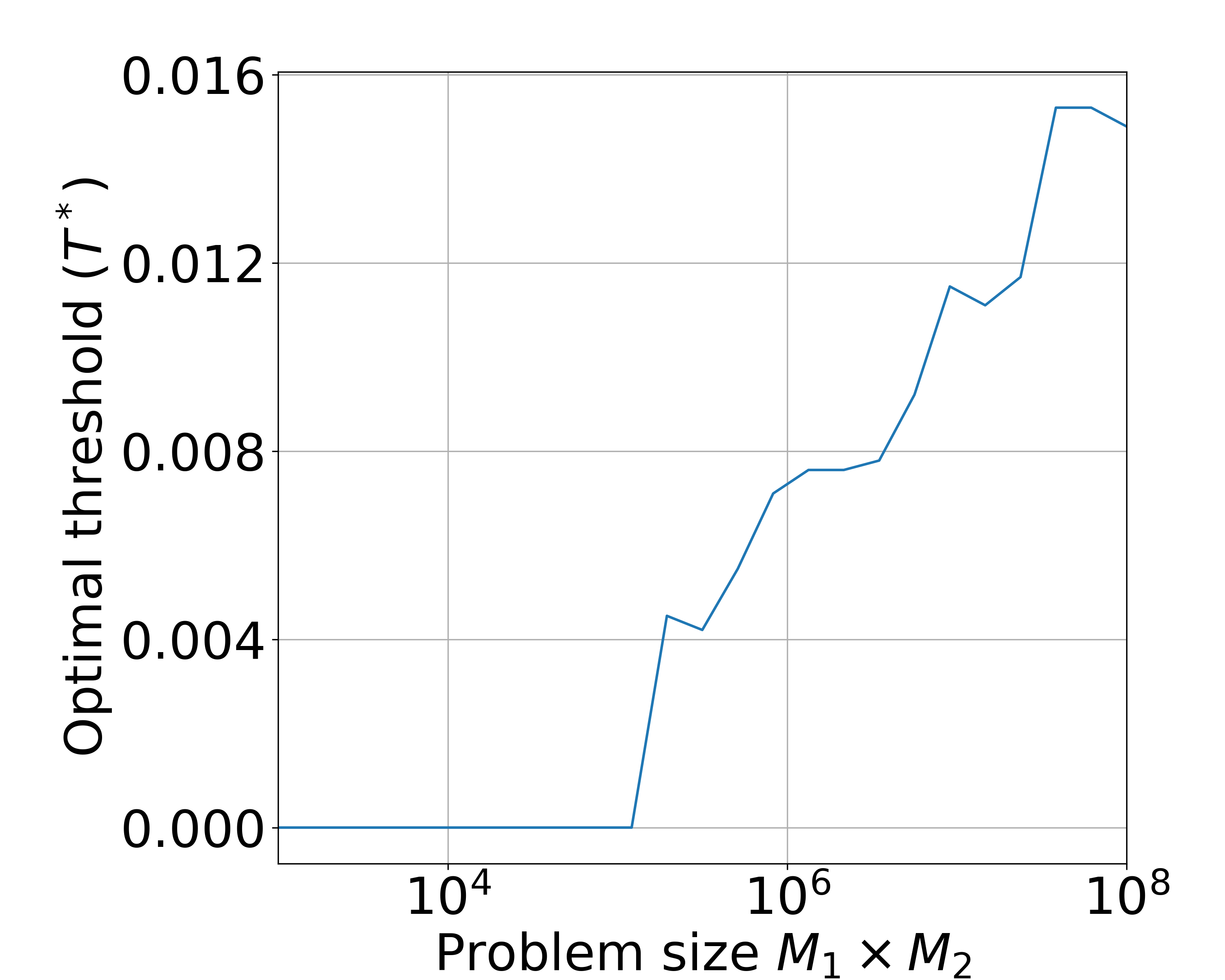}\label{fig:optThresh}}
\subfloat[Optimal sampling width ($A^*$).]{\includegraphics[width=0.35\linewidth]{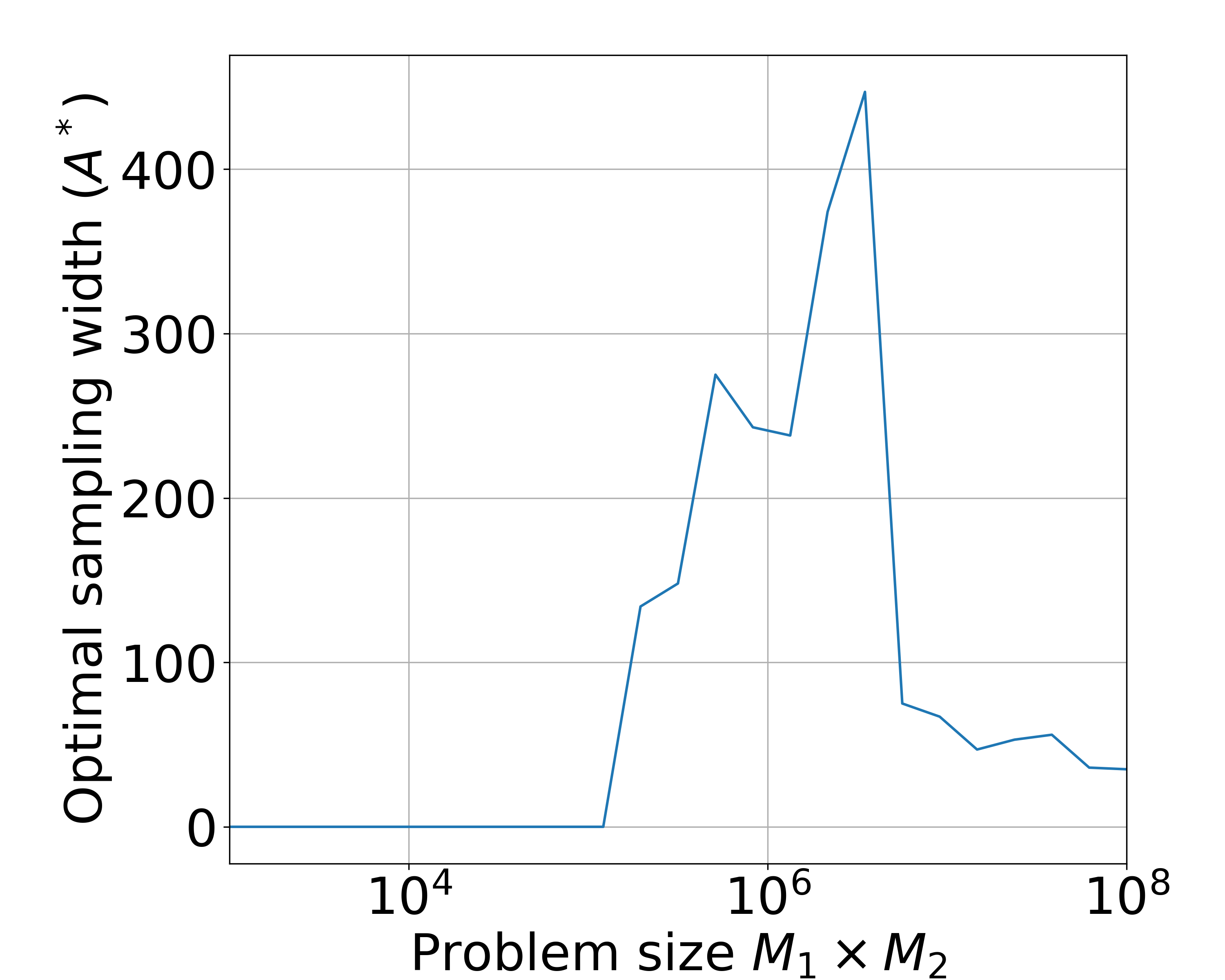}  \label{fig:optA}}  
\caption{Threshold and sampling width of Bayesian optimization for $D_p=512$, $B=4$, and $F=2$.}
\label{fig:hyperparams}
\end{figure}

\subsection{Hyperparameter optimization}
This section explains the methodology for finding optimal \name hyperparameters to achieve high accuracy and fast convergence.
The optimal configuration is denoted by $\mathbf{c^\star}=(T^\star, A^\star)$, corresponding to the optimal threshold and sampling width.
As an automatic hyperparameter search method, we employ Bayesian optimization\rebuttal{~\cite{snoek2012practical}}.
The loss function is defined as the error rate given by the percentage of incorrect factorizations out of 512 randomly selected product vectors.
To put a strong emphasis on fast convergence, we reduced the maximal number of the iterations to $N' = 0.05 N$ for all Bayesian optimization runs. 
The error rate is an unknown function of the hyperparameters, modeled as a Gaussian process with a radial basis function kernel.
Hyperparameter sampling is done using the expected improvement acquisition function.
For each problem ($F$, $D_p$, $\prod_{f=1}^{F} M_f$, $B$), we run five separate hyperparameter searches, each of which tests 200 different hyperparameter combinations restricted to the domains $A \in [0,M_f]$ and $T \in [0,1]$.
Finally, we select the hyperparamters with the lowest error rate at the default maximum number of iterations ($N$). 

Fig.~\ref{fig:hyperparams} shows the resulting threshold ($T$) and sampling width ($A$) over various problem sizes for $D_p=512$, $B=4$, and $F=2$. 
For a range of problem sizes ($10^3$--$10^5$), \name does not require the threshold and sampling dynamics: it sets the threshold and the sampling width to 0.
For larger problem sizes ($>$$10^5$), the threshold $T$ grows with the problem size. 
This can be explained by the fact that querying larger codebooks is likely to activate more codevectors, which will be bundled. 
As such, we expect a higher interference between likely incorrect low-similarity solutions and promising high-similarity solutions. 
A higher threshold effectively reduces the number of bundled vectors, reducing interference. 
Similarly, the sampling width ($A$) grows until a problem size of 4,000,000, where it sharply declines.
The sharp decline was observed for all investigated problem settings ($F$, $D_p$, $B$) and might stem from the limited bundling capacity.

\subsection{Experimental setup}
\begin{figure*}
    \centering
    \includegraphics[width=\textwidth]{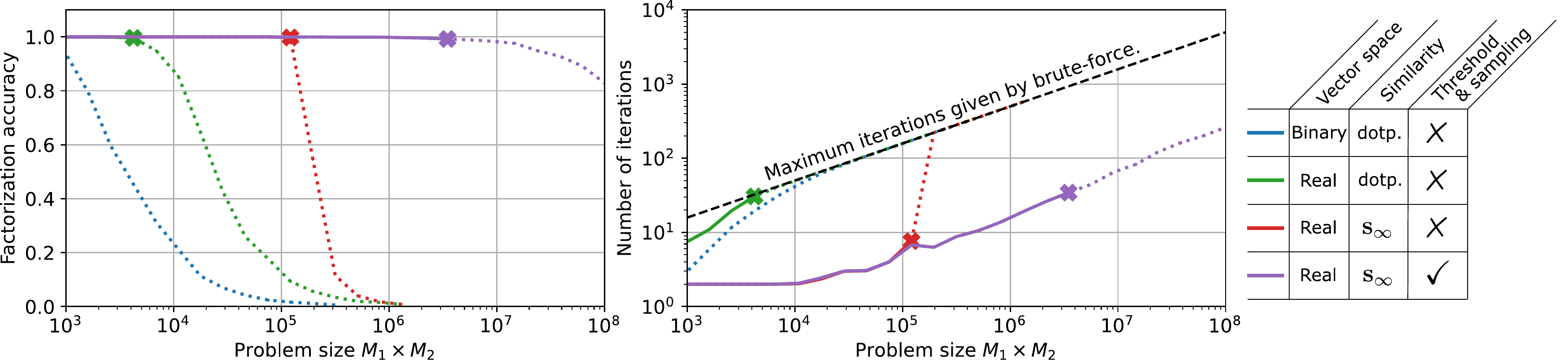}
    \fontsize{7}{9}
    \selectfont
    % \vspace*{-8mm}
    \caption{Factorization accuracy (left) and number of iterations (right) of various \name configurations on synthetic (i.e., exact) product vectors for different problem sizes ($\prod_{f=1}^{F} M_f$). 
    We set $D_p=512$, $F=2$, and $B=4$. 
    The maximum operational capacity is marked with a cross. Problem sizes exceeding the operational capacity are marked with dashed lines which face an accuracy lower than 99\%. \name configured with binary SBC operations (in blue) cannot solve any of the displayed problem sizes at the required accuracy. 
    }
  \label{fig:factorizer-main-results}
\end{figure*}
We evaluate the performance of our novel \name on randomly selected synthetic product vectors.
For each problem ($F$, $D_p$, $\prod_{f=1}^{F} M_f$, $B$), we assess the factorization accuracy and the number of iterations by averaging over 5000 experiments. In each experiment, we randomly select one vector from each of the $F$-many codebooks, bind the selected vectors together to form a product vector, then use the product vector as the input into \name. In this section, the queries are always binary SBCs. In contrast, the representations inside \name (i.e., factor estimates at any step of the decoding loop) do not have this restriction imposed upon them unless specified otherwise.
The operational capacity is defined as the largest problem size for which \name achieves an accuracy higher than $99\%$~\cite{Resonator2}.
\subsection{Comparative results}\label{subsec:fac_results}
Fig.~\ref{fig:factorizer-main-results} compares the accuracy (left) and the number of iterations (right) of various \name configurations with $D_p=512$ and $F=2$. 
Dotted lines indicate a less than $99\%$ factorization accuracy. 
Starting with binary SBC vectors and the dot-product similarity metric, we can see that this \name configuration fails to solve any problem of size larger than $10^3$ accurately.
The operational capacity increases to $4.2\cdot10^3$ when relaxing the sparsity constraint of binary SBCs by allowing for \gls{code} representations inside \name. 
%$4200$.
%
However, the required iterations are still high, requiring almost as many searches as the brute-force approach.
The introduction of the $\ell_\infty$-based similarity increases the operational capacity by more than an order of magnitude. 
We can also notice a drastic reduction in the number of iterations necessary for converging to the correct solution. 
For problem sizes up to $10^{4}$, \name needs only $2$ iterations to converge, the minimum possible number of iterations to detect convergence reliably.
However, as the problem size goes beyond $1.2\cdot10^5$, \name encounters limit cycles and spurious fixed points, hindering its convergence to the correct solution~ \cite{Resonator2}. 
To this end, we introduce the threshold nonlinearity coupled with conditional random sampling. 
With these new dynamics, \name further increases the operational capacity by over an order of magnitude to $5\cdot 10^6$. 

\begin{figure*}
    \includegraphics[width=\textwidth]{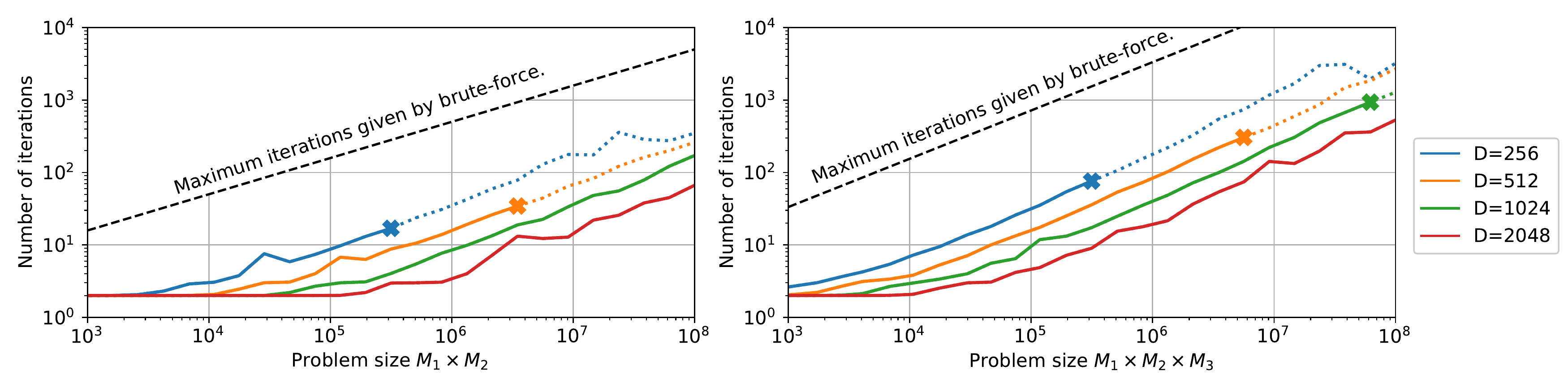}
    \fontsize{7}{9}
    \vspace*{-8mm}
    \selectfont
    \caption{Effect of the dimension $D_p$ on the number of iterations for \name with $B=4$, $F=2$ (left) and $F=3$ (right).
    }
  \label{fig:figure3}
\end{figure*}

\begin{figure*}
    \includegraphics[width=.99\textwidth]{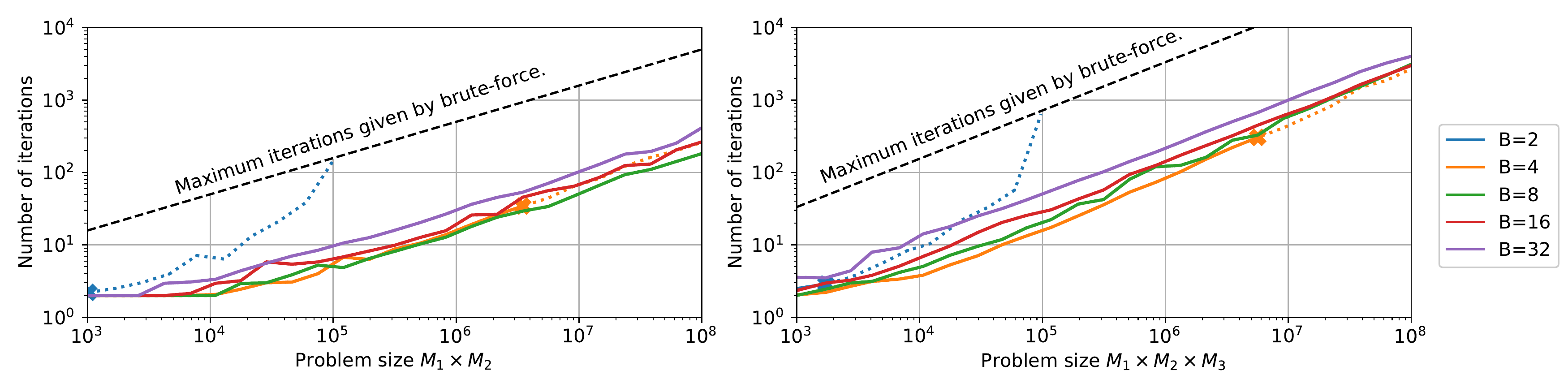}
    \fontsize{7}{9}
    \vspace*{-5mm}
    \selectfont
    \caption{Effect of the number of blocks $B$ on the number of iterations for \name with $D=512$, $F=2$ (left) and $F=3$ (right).
    }
  \label{fig:figure4}
\end{figure*}

Next, we analyze \name's decoding performance for a varying number of blocks ($B$), vector dimensions ($D_p$), and numbers of factors ($F$).
Fig.~\ref{fig:figure3} shows the number of iterations of \name for $F=2$ (left) and $F=3$ (right) factors when varying the vector dimension.
For both two and three factors, the operational capacity and convergence speed increase with growing vector dimensionality. 
As we move from two to three factors, the operational capacity remains approximately the same while the number of decoding iterations increases. 
However, an increase in the number of iterations does not directly lead to higher computational cost as each iteration requires fewer search operations ($F\cdot \sum_f M_f$) for larger $F$ due to the $F$-root dependence of $M_f$. 
For example, at $D_p=512$ and $\prod_{f=1}^{F} M_f=10^6$, \name at $F=2$ requires, on average, $15.73$ iterations corresponding to a total of $31,460$ searches, whereas at $F=3$ it requires, on average, $85.17$ iterations and $25,551$ searches. 

Fig.~\ref{fig:figure4} shows \name's performance for a fixed $D_p$ while varying the number of the blocks ($B$) for two and three factors. 
For a very small number of blocks ($B=2$), the operational capacity lies at approximately $10^3$ for both $F=2$ and $F=3$.
The operational capacity increases to around $5\cdot10^6$ when $B=4$. 
Further increasing the number of blocks ($B\geq8$) exhibits an operational capacity beyond $10^8$, the largest problem size we measured. 
The convergence speed peaks at $B=4$ and $B=8$ blocks, gradually decreasing as the vectors get denser.
Overall, the experimental results shown in Fig.~\ref{fig:figure3} and Fig.~\ref{fig:figure4} demonstrate the broad applicability of our \name: it accurately solves factorization problems within the computational constraints for a wide range of problem sizes, block sizes ($B\geq4$), and number of factors ($F\in\{2,3\}$).

Finally, we compare our \name with the state-of-the-art stochastic factorizer~\cite{langenegger2022imcfac} operating with dense bipolar vectors. 
We fix the problem size to $10^6$ and compare the number of iterations for three configurations according to those featured in \cite{langenegger2022imcfac}, namely $F=2$ with $D_p=1024$, $F=3$ with $D_p=1536$, and $F=4$ with $D_p=2048$. 
We configure our \name with $B=4$ blocks. 
Table \ref{tab:dense_vs_sparse} summarizes the results. 
Both factorizers achieve $>99\%$ accuracy across all configurations, whereby our \name requires up to 6$\times$ fewer iterations. 

\begin{table}[]
\caption{Comparison between stochastic factorizer~\cite{langenegger2022imcfac} and our \name at problem size $10^6$. }
\label{tab:dense_vs_sparse}
\centering
% \resizebox{0.7\linewidth}{!}{
\begin{threeparttable}
\begin{tabular}{ccrr}
\toprule
& & \multicolumn{2}{c}{Number of iterations}\\
\cmidrule(r){3-4}
$F$ & $D$ &  Dense bipolar~\cite{langenegger2022imcfac} & \name ($B=4$)\\
\cmidrule(r){1-2}\cmidrule(r){3-3}\cmidrule(r){4-4}
2 & 1024  & 68.47 & 11.16\\
3 & 1536  & 72.48 & 52.91\\
4 & 2048  & 157.17 & 89.05\\

\bottomrule
\end{tabular}
\end{threeparttable}
% }
\end{table}

\subsection{Ablation study}\label{subsec:bcf-ablation}
This section provides more insights into \name's two main hyperparameters: the threshold ($T$) and the sampling width ($A$). 

\paragraph{Effect of sampling width in an unconditional random sampler}
 
The sampling width ($A$) determines how many codevectors will be randomly sampled and bundled in case the thresholded similarity is an all-zero vector. 
Intuitively, we expect too low sampling widths to result in a slow walk over the space of possible solutions. 
Alternatively, suppose the sampling width is too large (e.g., larger than the bundling capacity). In that case, we expect high interference between the randomly sampled codevectors to hinder the accuracy and convergence speed due to the limited bundling capacity.

\begin{figure}
    \centering
    \includegraphics[width=0.6\linewidth]{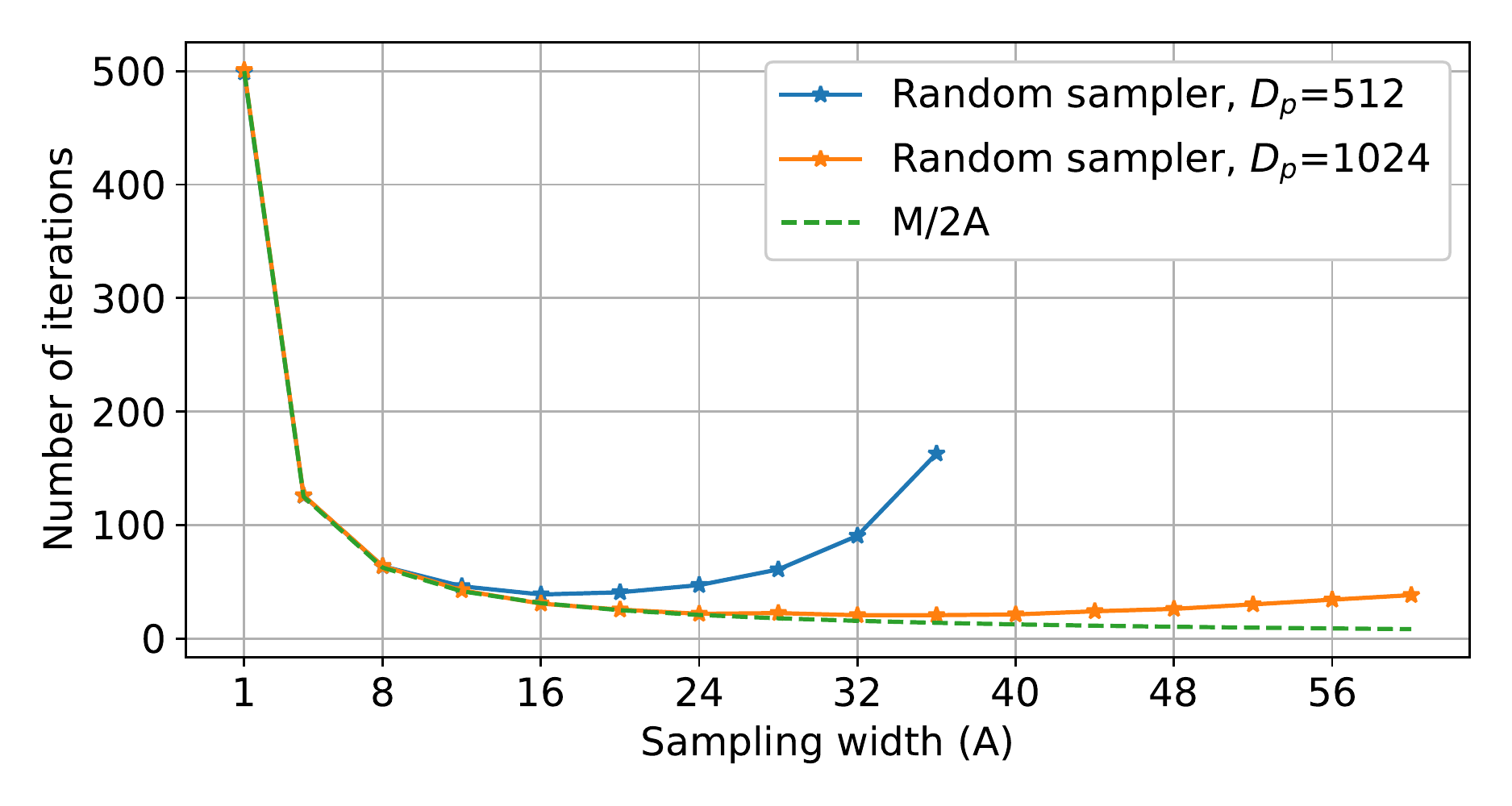}
    \vspace*{-3mm}
    \caption{Number of iterations when \name is configured as an unconditional random sampler with varying sampling width ($A$). We set $B=4$, $F=2$, and $M_{1}=M_{2}=M=1000$.
    }
  \label{fig:figureS2}
\end{figure}

To experimentally demonstrate this effect, we run \name with $F=2$ in an operational mode that corresponds to unconditional random vector sampling. 
Concretely, we fix the sampling width ($A$), generate the initial guess as a bundling of $A$-many randomly selected codevectors, and set both the sparsification threshold ($T$) and the convergence detection threshold ($T_c$) to the inverse sampling width ($1/A$). %
With this configuration, we expect \name to execute a random walk over the solution space with sampling width determining the number of solutions that are simultaneously evaluated. 
If the procedure samples the correct solution and the interference from sampled incorrect solutions is low, the correct factor triggers the threshold, and the factorization stops.
As a result, at every iteration, we test $M_f \cdot A^{F-1}$ combinations per factor~$f$. 
The expected number of iterations for this procedure is: 
\begin{align}
    \mathbb{E}[t] = \frac{\prod_{f=1}^{F} M_f}{(\sum_{f=1}^{F} M_f)\cdot A^{F-1}}, 
\end{align}
where the numerator reflects the overall problem size, and the denominator is the number of combinations the random sampler tests per iteration. 
With $F=2$ factors and codebooks of equal size $M_1=M_2=M$, the expected number of iterations equals $M/(2A)$.

Fig.~\ref{fig:figureS2} shows experimental results with this factorizer mode for $F=2$, $M=1000$, varying $D_p$ between 512 and 1024. 
The expected number of iterations corresponds to the expression $M/(2A)$. 
All presented configurations reach $>99\%$ accuracy, but in a different number of iterations.
For small sampling widths, the empirical results match the expectation. 
As the sampling width increases, the discrepancy between the empirical and theoretical results grows, more evidently at the smaller dimensions.
The discrepancy could stem from the bundling capacity, i.e., the number of retrievable elements, which decreases with a shrinking dimension. 

\paragraph{Effect of sampling width ($A$) in \name}

In this set of experiments, we do not restrict the threshold to be $1/A$.
Instead, we run a grid search for each sampling width over threshold values ($T$) in $[0,1]$ and use the optimal value in our benchmarks.

Table \ref{tab:varyA} shows how the accuracy and the number of iterations change as we vary the sampling width ($A$) in $\{10,50,100,200,500,1000\}$.
As expected, there is a sweet spot for sampling width, which lies at around 100 in this setting. Smaller values of sampling widths do not negatively impact the accuracy, but convergence speed does decrease. As we increase the sampling width beyond 100, the accuracy drops.
Moreover, with a fine-tuned threshold value, we can factorize product vectors notably faster (15.73 iterations) than when \name is run in the unconditional random sampling mode (38.87 iterations).

\begin{table}[]
\caption{\name performance when varying the sampling width ($A$). $D_p=512$, $F=2$, $M_{1}=M_{2}=1000$.}
\label{tab:varyA}
\centering
% \resizebox{0.7\linewidth}{!}{
\begin{threeparttable}
\begin{tabular}{rrrr}
\toprule
         \multicolumn{1}{c}{A}            & \multicolumn{1}{c}{$T^*$} & \multicolumn{1}{c}{Accuracy} & \multicolumn{1}{c}{Num. Iters.}\\
         \hline \\[-1.8ex]
                     %\cmidrule(r){1-1}\cmidrule(r){2-2}\cmidrule(r){3-3}\cmidrule(r){4-4}\cmidrule(r){5-5}\cmidrule(r){6-6}\cmidrule(r){7-9}
10 & 0.00602 & 99.4\% & 39.16\\
50 & 0.00641 & 99.4\% & 17.86\\
100 & 0.00641 & 99.4\% & 15.73\\
500 & 0.00722 & 98.6\% & 24.25\\
1000 & 0.00441 & 46.9\% & 276.78\\

\bottomrule
\end{tabular}
\end{threeparttable}
% }
\end{table}

\paragraph{Similarity metric}
\begin{figure}
\centering
\subfloat[$\ell_{\infty}$-based similarity.]{\includegraphics[width=0.35\linewidth]{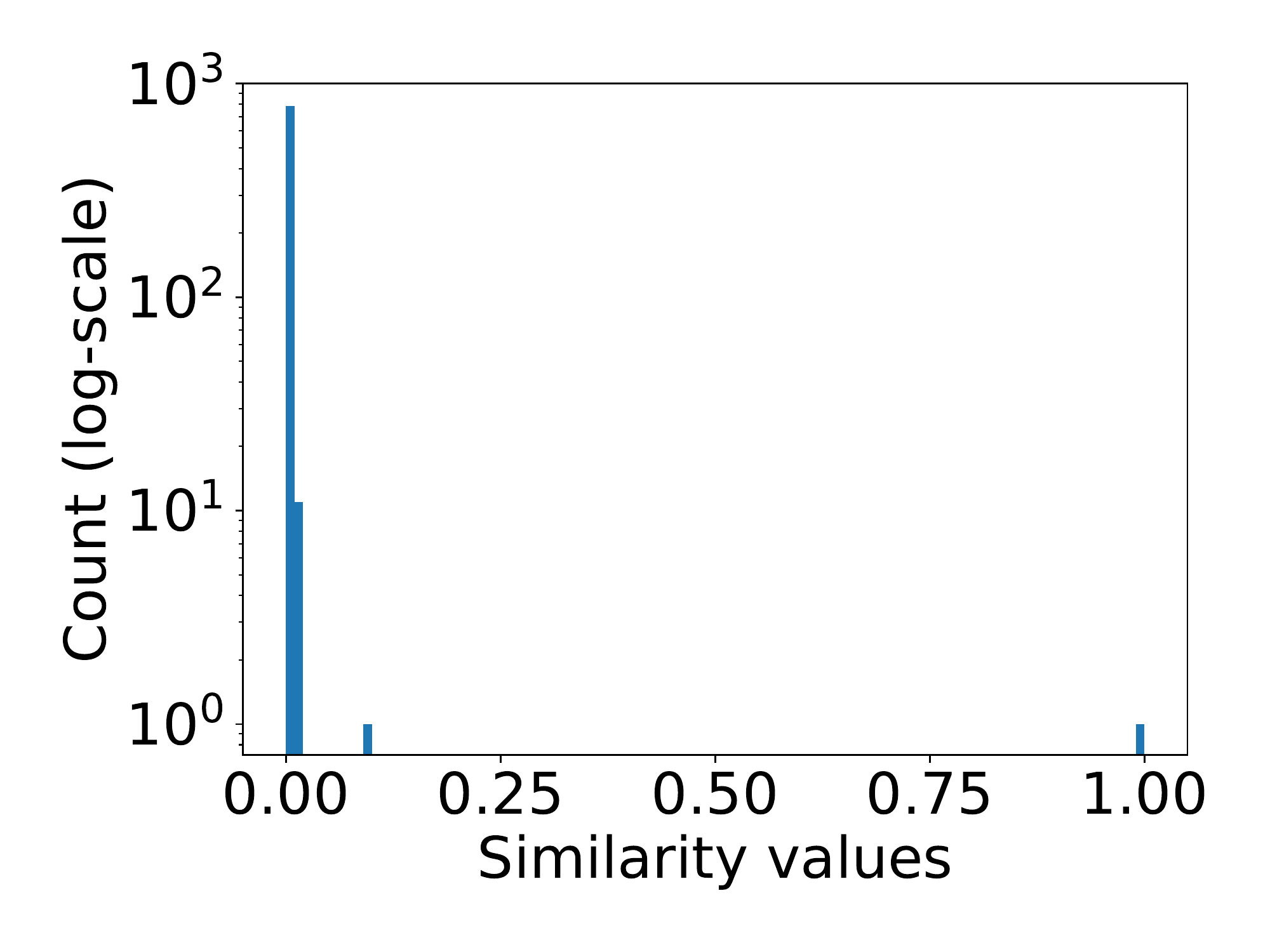} } \,
\subfloat[Dot-product similarity.]{\includegraphics[width=0.35\linewidth]{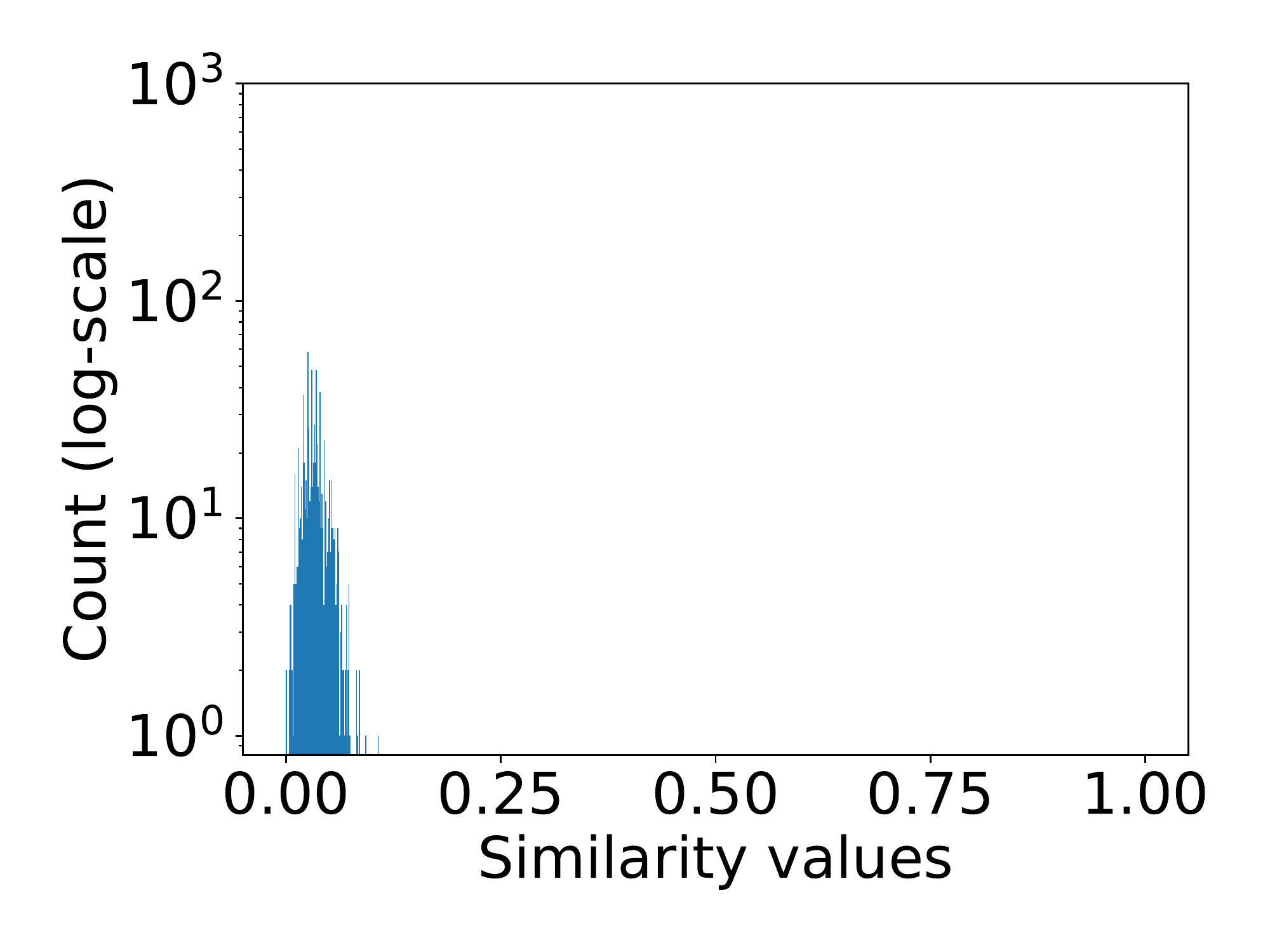} }
\caption{Log-scale histograms of $\ell_{\infty}$-based and dot-product similarities. \name with $D_p=512$, $B=4$, $F=2$, $M_1=M_2=200$, and $T=0$.}
\label{fig:figureS1}
\end{figure}

Here, we compare the $\ell_{\infty}$-based similarity with the dot-product similarity by considering the similarity distributions of the associative memory search inside \name. 
We select a problem size that cannot be solved by the dot-product similarity, but can be solved by the $\ell_{\infty}$-based similarity. 
The threshold nonlinearity and conditional random sampling are disabled. 
For $F=2$, $B=4$, and $D_{p}=512$, one such problem size is $4\cdot 10^4$.
We execute the decoding for two iterations and show the resulting histogram in Fig.~\ref{fig:figureS1}.
The $\ell_{\infty}$-based similarity tends to induce sparse activations: most similarities have a value of 0 and will have no effect on the weighted bundling of codevectors. 
Conversely, the dot-product similarity exhibits a wider distribution with almost no zero-valued similarities.
Finally, the $\ell_{\infty}$-based similarity found the correct solution (similarity value of 1), whereas the dot-product similarity did not.

\section{Part~II: Effective Replacement of Large FCLs with Block Code Factorizers}

\begin{figure}[]
    \centering

    \includegraphics[width=.5\linewidth]{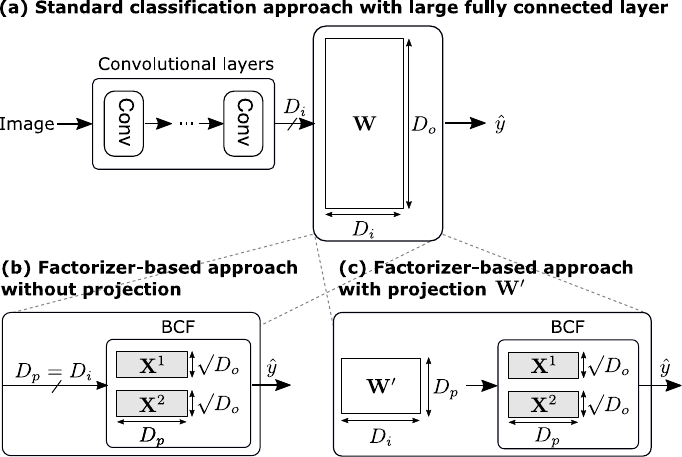}
    \caption{Replacement of a large FCL with our \name (b) without or (c) with a projection $\mathbf{W'}$. 
    }
  \label{fig:FCLreplacement}
\end{figure}

So far, we have applied our \name on synthetic (i.e., exact) product vectors. 
In this section, we present our second contribution, expanding the application of our \name to classification tasks in deep CNNs.
This is done by replacing the large final FCL in CNNs with our \name, as shown in Fig.~\ref{fig:FCLreplacement}. 
Instead of training $C$ hyperplanes for $C$ classes, embodied in the trainable weights $\mathbf{W}\in \Reals^{D_i \times D_o}$ of the FCL, where $C=D_o$, we represent the classes with a fixed binary SBC product space $\mathbf{P}\in \{0,1\}^{D_p \times D_o}$.
The product space requires only $B \cdot \sum_{f=1}^{F} M_f$ fixed integer values to be stored with the binary SBC offset notation accounting for 256 values on ImageNet-1K with $B=4$ and $M_1=M_2=32$. 
We provide two variants to interface the $D_i$-dimensional output features of the CNN's final convolutional layer with our \name, depicted in Fig.~\ref{fig:FCLreplacement}b and Fig.~\ref{fig:FCLreplacement}c. 
In the first variant, \name is directly interfaced with the CNN's output features; hence, the dimensionality becomes $D_p=D_i$. 
The second variant uses an intermediate, trainable projection $\mathbf{W'}\in \Reals^{D_i \times D_p}$, where $D_p \ll D_o$, % 
The number of parameters is notably reduced in both variants, by $D_o\cdot D_i$ without the projection, and by $(D_o-D_p)\cdot D_i$ with the projection. 

\subsection{Casting classification as a factorization problem}
  First, we describe how the classification problem can be transformed into a factorization problem. 
The codebooks and product space are naturally provided if a class is a combination of multiple attribute values.
For example, the RAVEN dataset contains different objects formed by a combination of shape, position, color, and size. 
Hence, we define four codebooks ($\mathbf{X}^1$, $\mathbf{X}^2$, $\mathbf{X}^3$, and $\mathbf{X}^4$) where the size of each codebook ($M_f$) corresponds to the number of values the individual attribute can have~\cite{hersche2022nvsa} (e.g., the codebook $\mathbf{X}^1$ representing five shapes has $M_1=5$ elements). 
The resulting product space is $\mathbf{P}=\mathbf{X}^1\oast\mathbf{X}^2\oast\mathbf{X}^3\oast\mathbf{X}^4$. 

% arbitrary classification problem
If no such semantic information is available, the codebooks and product space are chosen arbitrarily. 
When targeting two factors, we first define a product space $\mathbf{P}=\mathbf{X}^1 \oast \mathbf{X}^2$ that contains $M_1\cdot M_2$ unique quasi-orthogonal product vectors. 
The size of the product space is set to the number of classes $C$, such that each product vector in $\mathbf{P}$ can be assigned to a unique class. 
For example, for representing the $C=100$ classes in the CIFAR-100 dataset, we define a product space with size $100$ using two codebooks of size $M_1=M_2=\sqrt{C}=10$. 
Then, the product vector $\mathbf{p}_1:= \mathbf{x}^1_1\oast\mathbf{x}^2_1$ belongs to ``\textit{class 1}'' and  $\mathbf{p}_{100}:= \mathbf{x}^1_{10}\oast\mathbf{x}^2_{10}$ to ``\textit{class 100}''\footnote{If $\sqrt{C}$ is not an integer, we take the ceiling of $\sqrt{C}$.}.

\begin{figure}[t]
    \centering
    \includegraphics[width=0.5\linewidth]{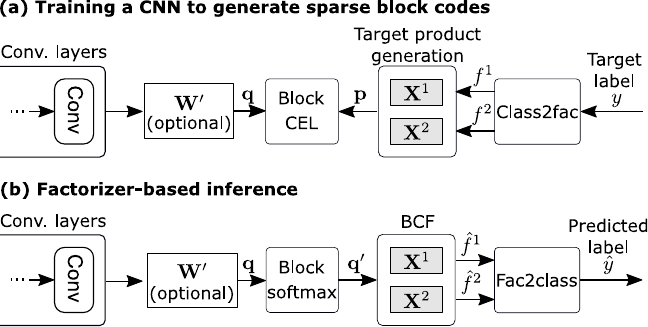}
    \caption{Training and inference with \name. 
    }
  \label{fig:FCLtraining}
\end{figure}

\subsection{Training CNNs with blockwise cross-entropy loss}
After defining the product space, we train a function $f_\theta$ (e.g., a CNN) with trainable parameters $\theta$ to map the input data (e.g., images) to the target product vectors of the corresponding classes. 
Fig.~\ref{fig:FCLtraining}a illustrates the training procedure.
Given a labeled training sample $(\mathbf{I},y)$ containing the input image $\mathbf{I}\in \Reals^{c \times h \times w}$ and the target label $y$, we first pass the image through the function $f_\theta$, yielding $\mathbf{q}=f_\theta(\mathbf{I})$. 
Next, we generate the target product vector by mapping the target label $y$ to the factor indices ($f^1$ and $f^2$) and forming the corresponding product $\mathbf{p} = \mathbf{x}^1_{f^1} \oast \mathbf{x}^2_{f^2}$. 

A typical loss function for binary sparse target vectors is the binary cross-entropy loss in connection with the sigmoid nonlinearity.  
However, we experienced a notable classification drop when using the binary cross-entropy loss; e.g., the accuracy of MobileNetV2 on the ImageNet-1K dataset dropped below 1\% when using this loss function. 
To this end, we propose a novel blockwise loss that computes the sum of per-block categorical cross-entropy loss (CEL). 
For each block $b$, we extract the $L$-dimensional block $\mathbf{q}_b:=\mathbf{q}[(b-1)L+1:bL]$ from the output features $\mathbf{q}$ and the target index $\dot{\mathbf{p}}[b]$. 
Then, the blockwise CEL is defined as: 
\begin{align}
    \mathcal{L}\left(\mathbf{q}, \dot{\mathbf{p}}, s\right) = \frac{1}{B }\sum_{b=1}^B \mathcal{L}_{\mathrm{CEL}}\left( s \cdot \mathbf{q}_b,\dot{\mathbf{p}}[b] \right),
\end{align}
where $\mathcal{L}_{\mathrm{CEL}}$ is the categorical CEL, which combines the softmax activation with the negative log-likelihood loss, and $s$ a trainable inverse softmax temperature for improved training~\cite{hoffer2018fix,RandomClassVec2021,scott2021mises}. 
The loss function $\mathcal{L}$ is minimized by batched stochastic gradient descent (SGD). 

\subsection{\name-based inference}
The \name-based inference is illustrated in Fig.~\ref{fig:FCLtraining}b. 
We pass a query image ($\mathbf{I}$) through the CNN, yielding the query product ($\mathbf{q}$), which can be interpreted as a ``noisy'' version of the ground-truth product vector $\mathbf{p}$. 
We then search for the product vector $\hat{\mathbf{p}} \in \mathbf{P}$ with the highest similarity to $\mathbf{q}$. 
One baseline is to compute the similarity between $\mathbf{q}$ and each product vector in $\mathbf{P}$ in a brute-force manner; however, this requires many similarity computations and the storage of each product vector in $\mathbf{P}$. 
Instead, we search for the closest product vector by factorizing the query product vector using \name, shown in Fig.~\ref{fig:FCLtraining}b. 
We pass the output of the CNN through a blockwise softmax function with an inverse softmax temperature $s_F$, which shapes and normalizes the blockwise distribution of the query vector.
The optimal inverse softmax temperature was found to be $s_F=1.5$, based on a grid search on the ImageNet-1K training set with MobileNetV2, and applied for all architectures.

\subsection{Experimental Setup}

\paragraph{Datasets} We evaluate our new method on three image classification benchmarks.\\ 
\textbf{ImageNet-1K.} The ImageNet-1K dataset~\cite{deng2009imagenet} is a large-scale image classification benchmark with colored images from $C=1000$ different classes. 
The training set contains over $1.2$\,M samples, and the validation set includes $50,000$ samples ($50$ per class), all with a resolution of $224 \times 224$.\\ 
\textbf{CIFAR-100.} The CIFAR-100 dataset~\cite{krizhevsky2009learning} contains colored images with a resolution of $32 \times 32$ from $C=100$ classes. 
The dataset provides 500 samples for training and 100 for testing per class.\\
\textbf{RAVEN.} The RAVEN dataset~\cite{Raven_19} contains Raven's progressive matrices tests with gray-scale images with a resolution of $160\times 160$.
A test provides 8 context and 8 answer panels, each containing objects with the following attributes:  position, type, size, and color. 
In this work, we exclusively target the recognition of single objects inside the panels.
Therefore, we extract all panels with single objects from the center, 2x2 grid, and 3x3 grid constellation.
The extraction gives us a dataset with $136,321$ panels for training, $45,144$ for validation, and $45,144$ for testing. 
We combine the positions from the different constellations, yielding 14 unique positions: 1 for the center, 4 for the 2x2 grid, and 9 for the 3x3 grid. 
Overall, this dataset contains $C=4200$ attribute combinations (14 positions $\times$ 5 types $\times$ 6 sizes $\times$10 colors). 

\paragraph{Architectures}
ShuffleNetV2~\cite{ShuffleNetV2_ECCV2018}, MobileNetV2~\cite{MobileNetV2_CVPR2018}, ResNet-18, and ResNet-50~\cite{he2016deep} serve as baseline architectures. 
In addition to our \name-based replacement approach, we evaluate each architecture with the bipolar dense resonator~\cite{Resonator1, Resonator2}, the Hadamard readout~\cite{hoffer2018fix}, and the Identity readout~\cite{qian2020doweneed}. 
For the FCL replacement strategies without the intermediate projection layer, we removed the nonlinearity and batch norm of the last convolutional layer of all CNN architectures. 
This notably improved the accuracy of all replacement strategies; e.g., the accuracy of MobileNetV2 with the Identity replacement improved from $60.28\%$ to $70.72\%$ when removing the batch norm and the ReLU6 of the last convolutional layer. 
All other architectural blocks remained the same, including shortcuts in the last block of the ResNet-18 and ResNet-50. 
To apply the Identity approach where $D_i \neq D_o$, we used an identity matrix that reads out the first $D_o$ elements of the output feature vector and ignores the remaining $D_i-D_o$ elements. 
We could not adjust the dimension $D_i$ since it would have required additional adaptations in the CNNs, e.g., a downsampling layer in the shortcut connection of ResNet-50. 

\paragraph{Training setup}
The CNN models are implemented in PyTorch (version 1.11.0) and trained and validated on a Linux machine using up to 4 NVIDIA Tesla V100 GPUs with 32\,GB memory.
We train all CNN architectures with SGD with architecture-specific hyperparameters, described in Appendix~\ref{appx:cnn-training}.
For each architecture, we use the same training configuration for the baseline and all replacement strategies (i.e., Hadamard, Identity, resonator networks, and our \name). 
We repeat the main experiments five times with a different random seed and report the average results and standard deviation to account for training variability.

%
% Results without projection
\subsection{Comparative Results}
Table~\ref{tab:noFC} compares the classification accuracy of the baseline with various replacement approaches without projection, namely Hadamard~\cite{hoffer2018fix}, Identity~\cite{qian2020doweneed}, bipolar dense~\cite{Resonator1}, and our \name.
On ImageNet-1K, \name reduces the total number of parameters of deep CNNs by 4.4\%--44.5\%\footnote{The relative parameter reduction depends on the size of the overall network and the final FCL, which we replace with our method.}, while maintaining a high accuracy within $2.39\%$ across the majority of architectures, with the only exception of ShuffleNetV2 showing $4.46\%$ accuracy drop due to its very large FCL accounting for 44.5\% of total parameters.
Our \name matches the brute-force accuracy within $0.44\%$ in all architectures while requiring only $5$--$7$ iterations on average, despite the query product vectors from the CNNs being ``noisy.''
Inspired by extremely fast convergence on the synthetic experiments (see Fig.~\ref{fig:factorizer-main-results}), we could show that \name can match the brute-force accuracy within $0.54\%$ when only allowing up to $N=3$ iterations (see Appendix~\ref{appx:n}).  
This does not hold for the bipolar dense resonators showing notable accuracy drop (up to $16.22\%$), compared to the brute-force search, despite allowing a high number of iterations ($N=150$) and conducting extensive hyperparameter tuning across various loss functions, including arcface~\cite{arcface_19} (see Appendix~\ref{appx:bipolar}). 

\begin{table*}[t]
\centering
\caption{Comparison of approaches which replace the final FCL without any projection layer ($D_i=D_p$). We report the average accuracy $\pm$ the standard deviation over five runs with different seeds for the baseline and our \glspl{code}.
}
\label{tab:noFC}
\resizebox{\linewidth}{!}{
\begin{threeparttable}
\begin{tabular}{lrccccrcccrrr}
\toprule
 &       &    &          \multicolumn{10}{c}{FCL replacement approach} \\
 \cmidrule(r){4-13}
 &       &    &          &                & \multicolumn{3}{c}{Bipolar dense} & \multicolumn{5}{c}{Our \name ($B=4$)}        \\
\cmidrule(r){6-8}\cmidrule(r){9-13}
\begin{tabular}[c]{@{}l@{}}Dataset/\\ architecture\end{tabular}  & ($D_i$, $D_o$) &\begin{tabular}[c]{@{}c@{}}Baseline \\ acc. \end{tabular} & \begin{tabular}[c]{@{}c@{}}Had. \\ acc.\end{tabular} & \begin{tabular}[c]{@{}c@{}}Id. \\ acc.\end{tabular} &\begin{tabular}[c]{@{}c@{}}BF \\ acc.\end{tabular}      & \begin{tabular}[c]{@{}c@{}}Res. \\ acc.$^*$\end{tabular} & \begin{tabular}[c]{@{}c@{}}Avg. \\ iter.$^*$\end{tabular}    & \begin{tabular}[c]{@{}c@{}}BF \\ acc.\end{tabular}  & \begin{tabular}[c]{@{}c@{}}Fac. \\ acc.\end{tabular}  & \begin{tabular}[c]{@{}c@{}}Avg. \\ iter.\end{tabular} & \begin{tabular}[c]{@{}l@{}}Param. \\saving$\uparrow$\end{tabular} & \begin{tabular}[c]{@{}c@{}}FCL\,comp. \\saving$\uparrow$\end{tabular} \\
\cmidrule(r){1-1}\cmidrule(r){2-2}\cmidrule(r){3-3}\cmidrule(r){4-4}\cmidrule(r){5-5}\cmidrule(r){6-6}\cmidrule(r){7-8}\cmidrule(r){9-9}\cmidrule(r){10-13}\morecmidrules\cmidrule(r){10-13}
\textbf{ImageNet-1K}     \\
ShuffleNetV2          & (1024, 1k) &  $69.22^{\pm0.20}$  &  $68.02$  &     $67.62$ & $66.17$  & $54.54$  & $150$ & $65.09^{\pm0.10}$ & $64.76^{\pm0.13}$  & $7$ &  $44.5\%$ & $55.2\%$\\
MobileNetV2           & (1280, 1k)  &  $71.57^{\pm0.13}$  &  $71.30$       &     $70.72$ & $70.64$ & $60.83$  & $150$ &  $70.00^{\pm0.07}$ & $69.76^{\pm0.13}$ & $6$ &  $37.6\%$ & $61.6\%$\\
ResNet-18             & (512, 1k) &  $70.39^{\pm0.11}$  &  N/A        &  N/A        &   $68.65$ & $54.17$  & $150$ &$68.44^{\pm0.08}$ & $68.00^{\pm0.07}$ & $7$ & $4.4\%$ & $55.2\%$\\
ResNet-50             & (2048, 1k)  &  $76.21^{\pm0.28}$  &  $75.30$       &  $74.65$  & $75.80$ & $67.98$ & $150$  & $76.34^{\pm0.04}$ & $76.25^{\pm0.07}$ & $5$ & $8.0\%$ & $68.0\%$\\
\cmidrule(r){1-1}\cmidrule(r){2-2}\cmidrule(r){3-3}\cmidrule(r){4-4}\cmidrule(r){5-5}\cmidrule(r){6-6}\cmidrule(r){7-8}\cmidrule(r){9-9}\cmidrule(r){10-13}\morecmidrules\cmidrule(r){10-13}
\textbf{CIFAR-100} \\
ResNet-18             & (512, 100)  &  $78.10^{\pm0.31}$  &  $77.21$  &  $76.56$    & $76.63$ & $71.15$ & $150$ & $77.31^{\pm0.15}$ & $77.19^{\pm0.17}$ & $2$ &  $0.5\%$ & $60.0\%$ \\
\cmidrule(r){1-1}\cmidrule(r){2-2}\cmidrule(r){3-3}\cmidrule(r){4-4}\cmidrule(r){5-5}\cmidrule(r){6-6}\cmidrule(r){7-8}\cmidrule(r){9-9}\cmidrule(r){10-13}\morecmidrules\cmidrule(r){10-13}
\textbf{RAVEN} \\
ResNet-18             & (512, 4.2k)  &  $99.88^{\pm0.01}$  &  N/A       & N/A       &   $99.89$  &  $94.92$ & $45$  &  $99.87^{\pm0.01}$  &  $99.82^{\pm0.02}$ & $16$  & $16.2\%$ & $86.7\%$ \\
\cmidrule(r){1-1}\cmidrule(r){2-2}\cmidrule(r){3-3}\cmidrule(r){4-4}\cmidrule(r){5-5}\cmidrule(r){6-6}\cmidrule(r){7-8}\cmidrule(r){9-9}\cmidrule(r){10-13}\morecmidrules\cmidrule(r){10-13}
p-value            & \multicolumn{1}{c}{--}  &  \multicolumn{1}{c}{--}  &  $0.125$       & $0.125$       &   $0.063$  &  $0.031$ & \multicolumn{1}{c}{--}  &  \multicolumn{1}{l}{$0.094$}  & \multicolumn{1}{l}{$0.063$}   & \multicolumn{1}{r}{--} & \multicolumn{1}{c}{--} & \multicolumn{1}{c}{--} \\
\bottomrule
\end{tabular}
\begin{tablenotes}\footnotesize
\item[] {Acc.= Accuracy (\%); N/A= Not applicable; Had.= Reproduced Hadamard~\cite{hoffer2018fix}; Id.= Reproduced Identity~\cite{qian2020doweneed}; BF= Brute-force; Res.= Resonator nets~\cite{Resonator1}; Fac.= Block code factorizer whereby it sets $F=2$ for ImageNet-1K and CIFAR-100, and $F=4$ for RAVEN.
p-value is determined by signed Wilcoxon test with respect to baseline accuracy.}
\item[]$^*$ Maximum number of iterations was increased to $N=150$ for better performance in the resonator nets that leads to 10$\times$ more operations than the brute-force search. 
\end{tablenotes}
\end{threeparttable}
}
\end{table*}
\begin{table*}[]
\centering
\caption{Classification accuracy when interfacing the last convolution layer with \name using a projection layer with $D_p=512$. 
}
\label{tab:withFC}
\resizebox{0.85\linewidth}{!}{
\begin{tabular}{lrrcccrrc}
\toprule
 & \multicolumn{3}{c}{Baseline} & \multicolumn{5}{c}{Our \name ($D_p=512$, $B=4$)}                                                                      \\
\cmidrule(r){2-4}\cmidrule(r){5-9}
  \begin{tabular}[c]{@{}l@{}}Dataset/\\ architecture\end{tabular}                            & ($D_i$, $D_o$) & \# Param.           & Acc.          & \begin{tabular}[c]{@{}c@{}}BF \\ acc.\end{tabular}    & \begin{tabular}[c]{@{}c@{}}Fac. \\ acc.\end{tabular}    & \begin{tabular}[c]{@{}c@{}}Avg. \\ iter.\end{tabular} & \begin{tabular}[c]{@{}c@{}}Param. \\ saving$\uparrow$\end{tabular}  & \begin{tabular}[c]{@{}c@{}}FCL\,comp. \\ saving$\uparrow$\end{tabular} \\
\cmidrule(r){1-1}\cmidrule(r){2-4}\cmidrule(r){5-5}\cmidrule(r){6-9}\morecmidrules\cmidrule(r){6-9}
\textbf{ImageNet-1K}     \\
ShuffleNetV2             & (1024, 1k) &  $2.3$\,M  &  $69.22^{\pm0.20}$  &  $68.67^{\pm0.11}$  &  $68.41^{\pm0.14}$ & $6$  & $21.7\%$  & $29.6\%$ \\
MobileNetV2              & (1280, 1k) & $3.4$\,M  &  $71.57^{\pm0.13}$  &  $71.69^{\pm0.11}$  &  $71.49^{\pm0.10}$ & $6$ & $18.4\%$ & $33.4\%$  \\
ResNet-18                & (512, 1k) & $11.5$\,M  &   $70.39^{\pm0.11}$   &  $69.57^{\pm0.17}$  &  $69.19^{\pm0.13}$ & $6$  & $2.2\%$ & $10.4\%$ \\
ResNet-50                & (2048, 1k) &$25.5$\,M  &  $76.21^{\pm0.28}$  &  $76.72^{\pm0.06}$  &  $76.56^{\pm0.08}$ & $5$ & $3.9\%$ & $40.8\%$  \\
\cmidrule(r){1-1}\cmidrule(r){2-4}\cmidrule(r){5-5}\cmidrule(r){6-9}\morecmidrules\cmidrule(r){6-9}
\textbf{RAVEN} \\
ResNet-18               & (512, 4.2k) & $13.3$\,M  &  $99.88^{\pm0.01}$  &  $99.88^{\pm0.00}$  &  $99.85^{\pm0.02}$ & $16$  & $14.2\%$ & $78.6\%$\\
\cmidrule(r){1-1}\cmidrule(r){2-4}\cmidrule(r){5-5}\cmidrule(r){6-9}\morecmidrules\cmidrule(r){6-9}
p-value              & \multicolumn{1}{c}{--} & \multicolumn{1}{c}{--}  &  \multicolumn{1}{c}{--}  &  \multicolumn{1}{l}{$0.465$}  &  \multicolumn{1}{l}{$0.313$} &  \multicolumn{1}{r}{--} & \multicolumn{1}{c}{--} &\multicolumn{1}{c}{--} \\
\bottomrule
\end{tabular}
}
\end{table*}

On CIFAR-100, \name matches the baseline within $0.91\%$ with only 2 average iterations; while on RAVEN, it requires a slightly higher number of iterations ($16$) due to the larger number of factors ($F=4$) and the asymmetric codebook sizes. 
Across all datasets and architectures, our \name reduces the large FCL's computational cost by $55.2$--$86.7\%$.

Considering the other FCL replacement approaches, Hadamard consistently outperforms Identity. 
However, both the memory and computation requirements of Hadamard are $\mathcal{O}(D_i \cdot D_o)$, while our \name reduces both to $\mathcal{O}(D_i \cdot \sqrt{D_o})$, as $F=2$ and $N=3$ are constant. 
Hence, Hadamard is ineffective for a large value of $D_o$. 
Moreover, Identity is only competitive when $D_i$ is within the range of $D_o$ (MobileNetV2 and ShuffleNetV2); for other combinations, either it is not applicable (ResNet-18 where $D_i <D_o$), or ineffective (ResNet-50 where $D_i > D_o$). 

We compare our approach to weight pruning techniques, which usually sparsify the weights in all layers, whereas we focus on the final FCL due to its dominance in compact networks. 
Such pruning can be similarly applied to earlier layers in addition to our method. 
Pruning the final FCL of a pretrained MobileNetV2 with iterative magnitude-based pruning~\cite{zhu2017prune} yields notable accuracy degradation as soon as more than 95\% of the weights are set to zero. 
In contrast, our method remains accurate (69.76\%) in high sparsity regimes (i.e., 99.98\% zero elements). 
See Appendix~\ref{appx:pruning} for more details.

% Further comparison on CIFAR100
Furthermore, we compare our results with~\cite{RandomClassVec2021}, which randomly initialize the final FCL and keep it fixed during training. 
On CIFAR-100 with ResNet-18, they could show that fixing the final FCL layer even slightly improves the accuracy compared to the trainable FCL ($75.9\%$ vs. $74.9\%$; see their Table~1). 
However, our \name-based approach outperforms their fixed FCL approach ($77.19\%$) while reducing the memory requirements and the FCL compute cost. 

% Results w/ projectionT
Table~\ref{tab:withFC} shows the performance of our \name when using the projection layer ($D_p=512$).
The projection layer improves the \name-based accuracy in all benchmarks, especially in cases where the \name replacement approach faced challenges (i.e., ShuffleNetV2). 
Overall, we reduce the total number of parameters by 2.2\%--21.7\% while maintaining the accuracy within $1.2\%$, compared to the baseline with trainable FCL.

\subsection{Ablation Study}
We give further insights into the \name-based classification by analyzing the effect of the number of blocks, the projection dimension, the number of factors, and the initialization of the CNN weights. 

\paragraph{Number of blocks $B$}
Table~\ref{tab:B} shows the brute-force and \name classification accuracy for block codes with different numbers of blocks. 
The brute-force accuracy degrades as the number of blocks ($B$) increases, particularly in networks where the final FCL is dominant (e.g., ShuffleNetV2). 
These experiments demonstrate that deep CNNs are well-matched with very sparse vectors (e.g., $B=4$), and motivated us to devise a \name that can factorize product vectors with such a low number of blocks. 
Our \name achieves an accuracy within $0.43\%$ of the brute-force accuracy for product vectors with $B=4$.
For a larger number of blocks ($B\geq8$), our \name matches the brute-force accuracy within $3.7\%$.
Note that \name's hyperparameters were exclusively tuned for $B=4$, and then applied for other blocks. 
Hence, \name for $B=\{8,16,32\}$ could be further improved by hyperparameter tuning.

\paragraph{Projection dimension $D_p$}
We varied the projection dimension ($D_p$) from 128 (high reduction) to 1000 (no reduction since $D_p=D_o$) for MobileNetV2 on ImageNet-1K. 
With an extremely low dimension ($D_p=128$), \name shows a $2.86\%$ accuracy drop compared to the baseline with trainable FCL while saving $32.8\%$ of the parameters.
When going to higher dimensions (e.g., $D_p=768$), \name even surpasses the baseline accuracy while saving $8.7\%$ of the parameters. 
See Appendix~\ref{appx:d}. 

\begin{table*}[t]
\caption{Classification accuracy (\%) on ImageNet-1K for the baseline and the block code-based replacement approaches ($F=2$) with different number of blocks ($B$).
Lower number of blocks ($B$) results in higher accuracy.}
\label{tab:B}
\resizebox{\linewidth}{!}{
\begin{tabular}{lrllllllll}
\toprule
                                                                  &    & \multicolumn{2}{c}{ShuffleNetV2} & \multicolumn{2}{c}{MobileNetV2} & \multicolumn{2}{c}{ResNet-18} & \multicolumn{2}{c}{ResNet-50} \\
\cmidrule(r){1-1}\cmidrule(r){2-2}\cmidrule(r){3-4}\cmidrule(r){5-6}\cmidrule(r){7-8}\cmidrule(r){9-10}
\begin{tabular}[c]{@{}l@{}}Classification\\ approach\end{tabular} & B  & \multicolumn{1}{c}{BF}             & \multicolumn{1}{c}{Fac.} & \multicolumn{1}{c}{BF}             & \multicolumn{1}{c}{Fac.}   & \multicolumn{1}{c}{BF}             & \multicolumn{1}{c}{Fac.}& \multicolumn{1}{c}{BF}             & \multicolumn{1}{c}{Fac.}         \\
\cmidrule(r){1-1}\cmidrule(r){2-2}\cmidrule(r){3-4}\cmidrule(r){5-6}\cmidrule(r){7-8}\cmidrule(r){9-10}
Baseline                                                     & --  & $69.22^{\pm0.20}$ & \multicolumn{1}{c}{--} & $71.57^{\pm0.13}$ &   \multicolumn{1}{c}{--} &  $70.39^{\pm0.11}$         &  \multicolumn{1}{c}{--}  &   $76.21^{\pm0.28}$     & \multicolumn{1}{c}{--}      \\
\cmidrule(r){1-1}\cmidrule(r){2-2}\cmidrule(r){3-4}\cmidrule(r){5-6}\cmidrule(r){7-8}\cmidrule(r){9-10}
\multirow{4}{*}{\glspl{code}} & 4  & $65.09^{\pm0.10}$ &    $64.76^{\pm0.13}$ & $70.00^{\pm0.07}$ & $69.76^{\pm0.13}$ & $68.43^{\pm0.08}$    &   $68.00^{\pm0.07}$ & $76.34^{\pm0.04}$ & $76.25^{\pm0.07}$ \\
   & 8  & $64.30$ & $63.65$ & $69.53$ & $69.20$ & $67.90$ & $67.76$ & $76.69$ & $76.48$\\
   & 16 & $63.22$ & $62.04$ & $69.34$ & $68.83$ & $67.02$ & $64.98$ & $76.52$ & $76.27$ \\
   & 32 & $61.96$ & $59.74$ & $69.12$ & $68.35$ & $65.09$ & $61.39$ & $76.08$ & $75.62$ \\
\bottomrule
\end{tabular}
}
\end{table*}

\begin{table}[t]
\caption{Loading a pretrained ResNet-18 model improves accuracy and training time. Classification accuracy (\%) on ImageNet-1K using \name with ResNet-18 (with projection $D_p=512$, $B=4$, $F=2$). 
}
\label{tab:pretrain}
% \resizebox{\linewidth}{!}{
\begin{tabular}{lcrll}
\toprule
         & \multicolumn{1}{c}{\begin{tabular}[c]{@{}c@{}}Pretrained\\ model\end{tabular}} & \multicolumn{1}{c}{\begin{tabular}[c]{@{}c@{}}Training\\ epochs\end{tabular}} & \multicolumn{1}{c}{\begin{tabular}[c]{@{}c@{}}BF\\ acc.\end{tabular}} & \multicolumn{1}{c}{\begin{tabular}[c]{@{}c@{}}Fac. \\ acc.\end{tabular}} \\
\cmidrule(r){1-1}\cmidrule(r){2-3}\cmidrule(r){4-5}
Baseline &   \xmark          & 100 & $70.39^{\pm0.11}$  & \multicolumn{1}{c}{--} \\
\glspl{code}     &   \xmark           & 100 & $69.57^{\pm0.17}$ & $69.10^{\pm0.14}$  \\
\cmidrule(r){1-1}\cmidrule(r){2-3}\cmidrule(r){4-5}
\multirow{3}{*}{\glspl{code}}      &   \cmark  & 50 & $68.50^{\pm0.14}$  & $68.05^{\pm0.08}$  \\
      &   \cmark  & 75 & $69.21^{\pm0.09}$  &  $68.83^{\pm0.07}$ \\
      &  \cmark   & 100 & $70.08^{\pm0.16}$  & $69.72^{\pm0.15}$       \\
\bottomrule
\end{tabular}
% }
\end{table}

\paragraph{Number of factors $F$}
So far, we have evaluated \name with two factors, each having codebooks of size $M_1=M_2=32$ on the ImageNet-1K dataset. 
We demonstrate \name's capability with $F=3$ codebooks of size $M_f=10$, which achieves similar accuracy while requiring a higher number of iterations (11 vs. 6) than $F=2$. 
However, since each iteration requires fewer search operations for $F=3$ ($30$ vs. $64$ searches), the overall saving in computational complexity of the FCL remains similar. 
See Appendix~\ref{appx:f}. 

\paragraph{Initialize ResNet-18 with pretrained weights}
Finally, we show that the training of \name-based CNNs can be improved by initializing their weights from a model that was pretrained on ImageNet-1K. 
Table~\ref{tab:pretrain} shows the positive impact of pretraining of ResNet-18 (with projection) on ImageNet-1K. 
The pretraining improves the accuracy of \name by $0.62\%$, compared to the random initialization, when keeping the number of epochs the same.  
Moreover, when reducing the number of training epochs to $75$ and $50$, \name still yielded accurate predictions ($68.83\%$ and $68.05\%$). 
This experiment shows that our \name-based method can be applied with reduced training cost if a pretrained model is available. 

\rebuttal{
\section{Discussion}

\name is a powerful tool to iteratively decode both synthetic and noisy product vectors by efficiently exploring the exponential search space using computation in superposition. 
As one viable application, this allowed us to effectively replace the final large FCL in CNNs, reducing the memory footprint and the computational complexity of the model while maintaining high accuracy. 
If the classes were a natural combination of multiple attribute values (e.g., the objects in RAVEN), we cast the classification as a factorization problem by defining codebooks per attribute, and their combination as vector products. 
In contrast, the codebooks and product space were chosen arbitrarily if the dataset did not provide such semantic information about the classes (e.g., ImageNet-1K or CIFAR-100). 
Instead of this random fixed assignment, one could use an optimized dynamic label-to-prototype vector assignment~\cite{saadabadi2024hyperspherical}. 
It would be also interesting if a product space could be \textit{learned}, e.g., by gradient descent, revealing the inherent structure and composition of the individual classes.
Besides, other applications may benefit from a structured product representation, e.g., representing birds as a product of attributes in the CUB dataset~\cite{wah2011caltech}. 
Indeed, high-dimensional distributed representations have already been proven to be helpful when representing classes as a superposition of attribute vectors in the zero-shot setting~\cite{ruffino2023zeroshot}. 
Representing the combination of attributes in a product space may further improve the decoding efficiency.
% and overall enhance reasoning of the final classification provided by the model~\cite{chen2019looks}. 

%
This work focuses on decoding single vector products; however, efficiently \textit{decoding superpositions} of vector products with our \name would be highly beneficial.
First, it would allow us to decode images with multiple objects (e.g., multiple shapes in an RPM panel on RAVEN). 
Second, it enables the replacement of arbitrary FCL in neural networks, which usually involve activating multiple neurons.
This limitation has been addressed in~\cite{hersche2023superposdec} albeit for dense codes, where a mixed decoding method efficiently extracts a set of vector products from their fixed-width superposition. 
The mixed decoding combines sequential and parallel decoding methods to mitigate the risk of noise amplification, and increases the number of vector products that can be successfully decoded.
However, the number of retrievable vector products in the superposition still needs to be higher to be able to replace arbitrary FCLs in neural networks.
Therefore, future work into advanced decoding techniques could improve this aspect of \name. 

Finally, our \name could enhance Transformer models~\cite{vaswani2017attention} on different fronts. 
First, large embedding tables are a bottleneck in Transformer-based recommendation systems, consuming up to 99.9\% of the memory~\cite{desai2022random}. 
Replacing the embedding tables with our fixed-width product space would reduce the memory footprint as well as the computational complexity in inference when leveraging our \name. 
Second, the internal feedforward layers in Transformer models could be replaced by BCF, specifically the first of the two FCLs which can be viewed as key retrieval in a key-value memory~\cite{FCLareKVmem2021}.
As elaborated in the previous paragraph, the number of decodable vector products in superposition is still limited. 
Hence, sparsely activated keys would be beneficial. 
It has been shown that these sparse activations can be found in the middle layers of Transformer models~\cite{ramsauer2021hopfield}. 
}

% Ablation study. 

\section{Conclusion}
We proposed an iterative factorizer for generalized sparse block codes.
Its codebooks are randomly distributed high-dimensional binary sparse block codes, whose number of blocks can be as low as four. 
The multiplicative binding among the codebooks forms a quasi-orthogonal product space that represents a large number of class categories, or combinations of attributes.
As a use-case for our factorizer, we also proposed a novel neural network architecture that replaces trainable parameters in an FCL (aka classifier) with our factorizer, whose reliable operation is verified by accurately classifying/disentangling noisy query vectors generated from various CNN architectures.
This quasi-orthogonal product space not only reduces the memory footprint and computational complexity of the networks working with it, but also can reserve a huge representation space to prevent future classes/combinations from coming into conflict with already assigned ones, thereby further promoting interoperability in a continual learning setting.

\section*{Acknowledgements}
This work is supported by the Swiss National Science foundation (SNF), grant no. 200800.

% The insight of reserving 

% Use \bibliography{yourbibfile} instead or the References section will not appear in your paper
\bibliographystyle{ios1}
\bibliography{bibliography}

\begin{thebibliography}{63}
% BibTex style file: ios1.bst, 2019-01-11
\ifx \bisbn   \undefined \def \bisbn  #1{ISBN #1}\fi
\ifx \binits  \undefined \def \binits#1{#1} \fi
\ifx \bauthor  \undefined \def \bauthor#1{#1} \fi
\ifx \bjtitle  \undefined \def \bjtitle#1{\textit{#1}}\fi
\ifx \batitle  \undefined \def \batitle#1{#1} \fi
\ifx \bctitle  \undefined \def \bctitle#1{#1} \fi
\ifx \bvolume  \undefined \def \bvolume#1{\textbf{#1}}\fi
\ifx \byear  \undefined \def \byear#1{#1} \fi
\ifx \bissue  \undefined \def \bissue#1{#1} \fi
\ifx \bfpage  \undefined \def \bfpage#1{#1} \fi
\ifx \blpage  \undefined \def \blpage #1{#1} \fi
\ifx \burl  \undefined \def \burl#1{#1} \fi
\ifx \doiurl  \undefined \def \doiurl#1{#1} \fi
\ifx \betal  \undefined \def \betal{et al.} \fi
\ifx \binstitute  \undefined \def \binstitute#1{#1} \fi
\ifx \beditor  \undefined \def \beditor#1{#1} \fi
\ifx \bpublisher  \undefined \def \bpublisher#1{#1} \fi
\ifx \bbtitle  \undefined \def \bbtitle#1{\textit{#1}} \fi
\ifx \bedition  \undefined \def \bedition#1{#1} \fi
\ifx \bseriesno  \undefined \def \bseriesno#1{#1} \fi
\ifx \blocation  \undefined \def \blocation#1{#1} \fi
\ifx \bsertitle  \undefined \def \bsertitle#1{#1} \fi
\ifx \bsnm \undefined \def \bsnm#1{#1} \fi
\ifx \bsuffix \undefined \def \bsuffix#1{#1} \fi
\ifx \bparticle \undefined \def \bparticle#1{#1} \fi
\ifx \barticle \undefined \def \barticle#1{#1} \fi
\ifx \botherref \undefined \def \botherref #1{#1} \fi
\ifx \url \undefined \def \url#1{#1} \fi
\ifx \bchapter \undefined \def \bchapter#1{#1} \fi
\ifx \bbook \undefined \def \bbook#1{#1} \fi
\ifx \bcomment \undefined \def \bcomment#1{#1} \fi
\ifx \oauthor \undefined \def \oauthor#1{#1} \fi
\ifx \citeauthoryear \undefined \def \citeauthoryear#1{#1} \fi
\ifx \texttildelow  \undefined \def \texttildelow{\symbol{126}} \fi
\def \endbibitem {}
\ifx \bconflocation  \undefined \def \bconflocation#1{#1} \fi

\bibitem{MAP_1998}
\begin{bchapter}
\bauthor{\binits{R.W.}~\bsnm{Gayler}},
\bctitle{Multiplicative Binding, Representation Operators \& Analogy},
in: \bbtitle{Advances in Analogy Research: Integration of Theory and Data from
  the Cognitive, Computational, and Neural Sciences},
\byear{1998}.
\end{bchapter}
\endbibitem

\bibitem{GaylerJackendoff2003}
\begin{bchapter}
\bauthor{\binits{R.W.}~\bsnm{Gayler}},
\bctitle{Vector Symbolic Architectures Answer {J}ackendoff's Challenges for
  Cognitive Neuroscience},
in: \bbtitle{{Joint International Conference on Cognitive Science
  (ICCS/ASCS)}},
\byear{2003},
pp.~\bfpage{133}--\blpage{138}.
\end{bchapter}
\endbibitem

\bibitem{PlateHolographic1995}
\begin{barticle}
\bauthor{\binits{T.A.}~\bsnm{Plate}},
\batitle{Holographic Reduced Representations},
\bjtitle{IEEE Transactions on Neural Networks}
\bvolume{6}(\bissue{3})
(\byear{1995}),
\bfpage{623}--\blpage{641}.
\end{barticle}
\endbibitem

\bibitem{PlateHolographic2003}
\begin{bbook}
\bauthor{\binits{T.A.}~\bsnm{Plate}},
\bbtitle{Holographic Reduced Representations: Distributed Representation for
  Cognitive Structures},
\bpublisher{{Center for the Study of Language and Information, Stanford}},
\byear{2003}.
\end{bbook}
\endbibitem

\bibitem{KanervaHyperdimensional2009}
\begin{barticle}
\bauthor{\binits{P.}~\bsnm{Kanerva}},
\batitle{Hyperdimensional Computing: An Introduction to Computing in
  Distributed Representation with High-Dimensional Random Vectors},
\bjtitle{Cognitive Computation}
\bvolume{1}(\bissue{2})
(\byear{2009}),
\bfpage{139}--\blpage{159}.
\end{barticle}
\endbibitem

\bibitem{LargePatternsGreatSymbols}
\begin{bchapter}
\bauthor{\binits{P.}~\bsnm{Kanerva}},
\bctitle{{Large Patterns Make Great Symbols: An Example of Learning from
  Example}},
in: \bbtitle{Hybrid Neural Systems},
\beditor{\binits{S.}~\bsnm{Wermter}} and
\beditor{\binits{R.}~\bsnm{Sun}}, eds,
\bpublisher{Springer Berlin Heidelberg},
\blocation{Berlin, Heidelberg},
\byear{2000},
pp.~\bfpage{194}--\blpage{203}.
ISBN \bisbn{978-3-540-46417-4}.
\end{bchapter}
\endbibitem

\bibitem{AnalRetrieval_2000}
\begin{botherref}
\oauthor{\binits{T.A.}~\bsnm{Plate}},
Analogy retrieval and processing with distributed vector representations,
\textit{Expert Systems}
(2000).
\end{botherref}
\endbibitem

\bibitem{AnalMapp_2009}
\begin{bchapter}
\bauthor{\binits{R.W.}~\bsnm{Gayler}} and
\bauthor{\binits{S.D.}~\bsnm{Levy}},
\bctitle{A distributed basis for analogical mapping},
in: \bbtitle{New Frontiers in Analogy Research: Proceedings of the Second
  International Analogy Conference-Analogy},
\byear{2009}.
\end{bchapter}
\endbibitem

\bibitem{RasmussenInductiveReasoning2011}
\begin{barticle}
\bauthor{\binits{D.}~\bsnm{Rasmussen}} and
\bauthor{\binits{C.}~\bsnm{Eliasmith}},
\batitle{A Neural Model of Rule Generation in Inductive Reasoning},
\bjtitle{Topics in Cognitive Science}
\bvolume{3}(\bissue{1})
(\byear{2011}),
\bfpage{140}--\blpage{153}.
\end{barticle}
\endbibitem

\bibitem{EmruliAnalogical2013}
\begin{bchapter}
\bauthor{\binits{B.}~\bsnm{Emruli}},
\bauthor{\binits{R.W.}~\bsnm{Gayler}} and
\bauthor{\binits{F.}~\bsnm{Sandin}},
\bctitle{Analogical Mapping and Inference with Binary Spatter Codes and Sparse
  Distributed Memory},
in: \bbtitle{International Joint Conference on Neural Networks (IJCNN)},
\byear{2013}.
\end{bchapter}
\endbibitem

\bibitem{hersche2022nvsa}
\begin{botherref}
\oauthor{\binits{M.}~\bsnm{Hersche}},
\oauthor{\binits{M.}~\bsnm{Zeqiri}},
\oauthor{\binits{L.}~\bsnm{Benini}},
\oauthor{\binits{A.}~\bsnm{Sebastian}} and
\oauthor{\binits{A.}~\bsnm{Rahimi}},
A neuro-vector-symbolic architecture for solving {R}aven’s progressive
  matrices,
\textit{Nature Machine Intelligence}
(2023),
1--13.
\end{botherref}
\endbibitem

\bibitem{hersche2023nvsa_nesy}
\begin{bchapter}
\bauthor{\binits{M.}~\bsnm{Hersche}},
\bauthor{\binits{M.}~\bsnm{Zeqiri}},
\bauthor{\binits{L.}~\bsnm{Benini}},
\bauthor{\binits{A.}~\bsnm{Sebastian}} and
\bauthor{\binits{A.}~\bsnm{Rahimi}},
\bctitle{{Solving Raven's Progressive Matrices via a Neuro-vector-symbolic
  Architecture}},
in: \bbtitle{Proceedings of the 17th International Workshop on Neural-Symbolic
  Learning and Reasoning (NeSy)},
\byear{2023}.
\end{bchapter}
\endbibitem

\bibitem{LedouxConcentration2001}
\begin{bbook}
\bauthor{\binits{M.}~\bsnm{Ledoux}},
\bbtitle{{The Concentration of Measure Phenomenon}},
\bpublisher{American Mathematical Society},
\byear{2001}.
\end{bbook}
\endbibitem

\bibitem{Resonator1}
\begin{barticle}
\bauthor{\binits{E.P.}~\bsnm{Frady}},
\bauthor{\binits{S.J.}~\bsnm{Kent}},
\bauthor{\binits{B.A.}~\bsnm{Olshausen}} and
\bauthor{\binits{F.T.}~\bsnm{Sommer}},
\batitle{Resonator Networks, 1: An Efficient Solution for Factoring
  High-Dimensional, Distributed Representations of Data Structures},
\bjtitle{Neural Computation}
\bvolume{32}(\bissue{12})
(\byear{2020}),
\bfpage{2311}--\blpage{2331}.
\end{barticle}
\endbibitem

\bibitem{Resonator2}
\begin{barticle}
\bauthor{\binits{S.J.}~\bsnm{Kent}},
\bauthor{\binits{E.P.}~\bsnm{Frady}},
\bauthor{\binits{F.T.}~\bsnm{Sommer}} and
\bauthor{\binits{B.A.}~\bsnm{Olshausen}},
\batitle{Resonator Networks, 2: Factorization Performance and Capacity Compared
  to Optimization-Based Methods},
\bjtitle{Neural Computation}
\bvolume{32}(\bissue{12})
(\byear{2020}),
\bfpage{2332}--\blpage{2388}.
\end{barticle}
\endbibitem

\bibitem{langenegger2022imcfac}
\begin{botherref}
\oauthor{\binits{J.}~\bsnm{Langenegger}},
\oauthor{\binits{G.}~\bsnm{Karunaratne}},
\oauthor{\binits{M.}~\bsnm{Hersche}},
\oauthor{\binits{L.}~\bsnm{Benini}},
\oauthor{\binits{A.}~\bsnm{Sebastian}} and
\oauthor{\binits{A.}~\bsnm{Rahimi}},
In-memory factorization of holographic perceptual representations,
\textit{Nature Nanotechnology}
(2023).
\end{botherref}
\endbibitem

\bibitem{RachkovskijBinding2001}
\begin{barticle}
\bauthor{\binits{D.A.}~\bsnm{Rachkovskij}} and
\bauthor{\binits{E.M.}~\bsnm{Kussul}},
\batitle{Binding and normalization of binary sparse distributed representations
  by context-dependent thinning},
\bjtitle{Neural Computation}
\bvolume{13}(\bissue{2})
(\byear{2001}),
\bfpage{411}--\blpage{452}.
\end{barticle}
\endbibitem

\bibitem{laiho2015sparse}
\begin{bchapter}
\bauthor{\binits{M.}~\bsnm{Laiho}},
\bauthor{\binits{J.H.}~\bsnm{Poikonen}},
\bauthor{\binits{P.}~\bsnm{Kanerva}} and
\bauthor{\binits{E.}~\bsnm{Lehtonen}},
\bctitle{High-dimensional computing with sparse vectors},
in: \bbtitle{2015 IEEE Biomedical Circuits and Systems Conference (BioCAS)},
\byear{2015}.
\end{bchapter}
\endbibitem

\bibitem{FradySDR2021}
\begin{botherref}
\oauthor{\binits{E.P.}~\bsnm{Frady}},
\oauthor{\binits{D.}~\bsnm{Kleyko}} and
\oauthor{\binits{F.T.}~\bsnm{Sommer}},
Variable Binding for Sparse Distributed Representations: Theory and
  Applications,
\textit{IEEE Transactions on Neural Networks and Learning Systems}
(2021).
\end{botherref}
\endbibitem

\bibitem{gripon2011sparse}
\begin{barticle}
\bauthor{\binits{V.}~\bsnm{Gripon}} and
\bauthor{\binits{C.}~\bsnm{Berrou}},
\batitle{Sparse neural networks with large learning diversity},
\bjtitle{IEEE Transactions on Neural Networks}
\bvolume{22}(\bissue{7})
(\byear{2011}),
\bfpage{1087}--\blpage{1096}.
\end{barticle}
\endbibitem

\bibitem{knoblauch2020iterative}
\begin{barticle}
\bauthor{\binits{A.}~\bsnm{Knoblauch}} and
\bauthor{\binits{G.}~\bsnm{Palm}},
\batitle{Iterative retrieval and block coding in autoassociative and
  heteroassociative memory},
\bjtitle{Neural Computation}
\bvolume{32}(\bissue{1})
(\byear{2020}),
\bfpage{205}--\blpage{260}.
\end{barticle}
\endbibitem

\bibitem{schlegel2022comparison}
\begin{barticle}
\bauthor{\binits{K.}~\bsnm{Schlegel}},
\bauthor{\binits{P.}~\bsnm{Neubert}} and
\bauthor{\binits{P.}~\bsnm{Protzel}},
\batitle{A comparison of vector symbolic architectures},
\bjtitle{Artificial Intelligence Review}
\bvolume{55}(\bissue{6})
(\byear{2022}),
\bfpage{4523}--\blpage{4555}.
\end{barticle}
\endbibitem

\bibitem{MasseFruitFly2009}
\begin{botherref}
\oauthor{\binits{N.Y.}~\bsnm{Masse}},
\oauthor{\binits{G.C.}~\bsnm{Turner}} and
\oauthor{\binits{G.S.X.E.}~\bsnm{Jefferis}},
Olfactory Information Processing in Drosophila,
\textit{Current Biology}
\textbf{19}(16)
(2009).
\end{botherref}
\endbibitem

\bibitem{HTM2017}
\begin{botherref}
\oauthor{\binits{Y.}~\bsnm{Cui}},
\oauthor{\binits{S.}~\bsnm{Ahmad}} and
\oauthor{\binits{J.}~\bsnm{Hawkins}},
The HTM Spatial Pooler—A Neocortical Algorithm for Online Sparse Distributed
  Coding,
\textit{Frontiers in Computational Neuroscience}
\textbf{11}
(2017).
\end{botherref}
\endbibitem

\bibitem{WILLSHAW1969}
\begin{barticle}
\bauthor{\binits{D.J.}~\bsnm{Willshaw}},
\bauthor{\binits{O.P.}~\bsnm{Buneman}} and
\bauthor{\binits{H.C.}~\bsnm{Longuet-Higgins}},
\batitle{Non-Holographic Associative Memory},
\bjtitle{Nature}
\bvolume{222}(\bissue{5197})
(\byear{1969}),
\bfpage{960}--\blpage{962}.
\end{barticle}
\endbibitem

\bibitem{OlshausenSparseAct1996}
\begin{barticle}
\bauthor{\binits{B.A.}~\bsnm{Olshausen}} and
\bauthor{\binits{D.J.}~\bsnm{Field}},
\batitle{Natural image statistics and efficient coding},
\bjtitle{Network: Computation in Neural Systems}
\bvolume{7}(\bissue{2})
(\byear{1996}),
\bfpage{333}--\blpage{339}.
\end{barticle}
\endbibitem

\bibitem{Bent2022SDRTrueNorth}
\begin{bchapter}
\bauthor{\binits{G.}~\bsnm{Bent}},
\bauthor{\binits{C.}~\bsnm{Simpkin}},
\bauthor{\binits{Y.}~\bsnm{Li}} and
\bauthor{\binits{A.}~\bsnm{Preece}},
\bctitle{Hyperdimensional computing using time-to-spike neuromorphic circuits},
in: \bbtitle{International Joint Conference on Neural Networks (IJCNN)},
\byear{2022}.
\end{bchapter}
\endbibitem

\bibitem{Renner2022SDRBindingSpike}
\begin{bchapter}
\bauthor{\binits{A.}~\bsnm{Renner}},
\bauthor{\binits{Y.}~\bsnm{Sandamirskaya}},
\bauthor{\binits{F.T.}~\bsnm{Sommer}} and
\bauthor{\binits{E.P.}~\bsnm{Frady}},
\bctitle{Sparse Vector Binding on Spiking Neuromorphic Hardware Using Synaptic
  Delays},
in: \bbtitle{International Conference on Neuromorphic Systems (ICONS)},
\byear{2022}.
\end{bchapter}
\endbibitem

\bibitem{ProdKey_NIPS19}
\begin{bchapter}
\bauthor{\binits{G.}~\bsnm{Lample}},
\bauthor{\binits{A.}~\bsnm{Sablayrolles}},
\bauthor{\binits{M.A.}~\bsnm{Ranzato}},
\bauthor{\binits{L.}~\bsnm{Denoyer}} and
\bauthor{\binits{H.}~\bsnm{Jegou}},
\bctitle{Large Memory Layers with Product Keys},
in: \bbtitle{Advances in Neural Information Processing Systems (NeurIPS)},
\byear{2019}.
\end{bchapter}
\endbibitem

\bibitem{FCLareKVmem2021}
\begin{bchapter}
\bauthor{\binits{M.}~\bsnm{Geva}},
\bauthor{\binits{R.}~\bsnm{Schuster}},
\bauthor{\binits{J.}~\bsnm{Berant}} and
\bauthor{\binits{O.}~\bsnm{Levy}},
\bctitle{Transformer Feed-Forward Layers Are Key-Value Memories},
in: \bbtitle{Conference on Empirical Methods in Natural Language Processing},
\byear{2021}.
\end{bchapter}
\endbibitem

\bibitem{XML-CNN2017}
\begin{bchapter}
\bauthor{\binits{J.}~\bsnm{Liu}},
\bauthor{\binits{W.-C.}~\bsnm{Chang}},
\bauthor{\binits{Y.}~\bsnm{Wu}} and
\bauthor{\binits{Y.}~\bsnm{Yang}},
\bctitle{Deep Learning for Extreme Multi-Label Text Classification},
in: \bbtitle{International ACM SIGIR Conference on Research and Development in
  Information Retrieval},
\byear{2017}.
\end{bchapter}
\endbibitem

\bibitem{ganesan2021learning}
\begin{bchapter}
\bauthor{\binits{A.}~\bsnm{Ganesan}},
\bauthor{\binits{H.}~\bsnm{Gao}},
\bauthor{\binits{S.}~\bsnm{Gandhi}},
\bauthor{\binits{E.}~\bsnm{Raff}},
\bauthor{\binits{T.}~\bsnm{Oates}},
\bauthor{\binits{J.}~\bsnm{Holt}} and
\bauthor{\binits{M.}~\bsnm{McLean}},
\bctitle{Learning with Holographic Reduced Representations},
in: \bbtitle{Advances in Neural Information Processing Systems (NeurIPS)},
\byear{2021}.
\end{bchapter}
\endbibitem

\bibitem{qian2020doweneed}
\begin{botherref}
\oauthor{\binits{Z.}~\bsnm{Qian}},
\oauthor{\binits{T.L.}~\bsnm{Hayes}},
\oauthor{\binits{K.}~\bsnm{Kafle}} and
\oauthor{\binits{C.}~\bsnm{Kanan}},
Do we need fully connected output layers in convolutional networks?,
\textit{arXiv preprint arXiv:2004.13587}
(2020).
\end{botherref}
\endbibitem

\bibitem{Raven_19}
\begin{bchapter}
\bauthor{\binits{C.}~\bsnm{Zhang}},
\bauthor{\binits{F.}~\bsnm{Gao}},
\bauthor{\binits{B.}~\bsnm{Jia}},
\bauthor{\binits{Y.}~\bsnm{Zhu}} and
\bauthor{\binits{S.-C.}~\bsnm{Zhu}},
\bctitle{RAVEN: A Dataset for Relational and Analogical Visual REasoNing},
in: \bbtitle{Proceedings of the IEEE Conference on Computer Vision and Pattern
  Recognition (CVPR)},
\byear{2019}.
\end{bchapter}
\endbibitem

\bibitem{deng2009imagenet}
\begin{bchapter}
\bauthor{\binits{J.}~\bsnm{Deng}},
\bauthor{\binits{W.}~\bsnm{Dong}},
\bauthor{\binits{R.}~\bsnm{Socher}},
\bauthor{\binits{L.-J.}~\bsnm{Li}},
\bauthor{\binits{K.}~\bsnm{Li}} and
\bauthor{\binits{L.}~\bsnm{Fei-Fei}},
\bctitle{Imagenet: A large-scale hierarchical image database},
in: \bbtitle{IEEE Conference on Computer Vision and Pattern Recognition
  (CVPR)},
\byear{2009},
pp.~\bfpage{248}--\blpage{255}.
\end{bchapter}
\endbibitem

\bibitem{HDCSurvey_PI}
\begin{botherref}
\oauthor{\binits{D.}~\bsnm{Kleyko}},
\oauthor{\binits{D.A.}~\bsnm{Rachkovskij}},
\oauthor{\binits{E.}~\bsnm{Osipov}} and
\oauthor{\binits{A.}~\bsnm{Rahimi}},
A Survey on Hyperdimensional Computing Aka Vector Symbolic Architectures, Part
  I: Models and Data Transformations,
\textit{ACM Comput. Surv.}
(2022).
\end{botherref}
\endbibitem

\bibitem{BinDensity_2018}
\begin{botherref}
\oauthor{\binits{D.}~\bsnm{Kleyko}},
\oauthor{\binits{A.}~\bsnm{Rahimi}},
\oauthor{\binits{D.A.}~\bsnm{Rachkovskij}},
\oauthor{\binits{E.}~\bsnm{Osipov}} and
\oauthor{\binits{J.M.}~\bsnm{Rabaey}},
Classification and Recall With Binary Hyperdimensional Computing: Tradeoffs in
  Choice of Density and Mapping Characteristics,
\textit{IEEE Transactions on Neural Networks and Learning Systems}
\textbf{29}(12)
(2018).
\end{botherref}
\endbibitem

\bibitem{renner2022neuromorphic}
\begin{botherref}
\oauthor{\binits{A.}~\bsnm{Renner}},
\oauthor{\binits{L.}~\bsnm{Supic}},
\oauthor{\binits{A.}~\bsnm{Danielescu}},
\oauthor{\binits{G.}~\bsnm{Indiveri}},
\oauthor{\binits{B.A.}~\bsnm{Olshausen}},
\oauthor{\binits{Y.}~\bsnm{Sandamirskaya}},
\oauthor{\binits{F.T.}~\bsnm{Sommer}} and
\oauthor{\binits{E.P.}~\bsnm{Frady}},
Neuromorphic visual scene understanding with resonator networks,
\textit{arXiv preprint arXiv:2208.12880}
(2022).
\end{botherref}
\endbibitem

\bibitem{kymn_compositional_2024}
\begin{botherref}
\oauthor{\binits{C.J.}~\bsnm{Kymn}},
\oauthor{\binits{S.}~\bsnm{Mazelet}},
\oauthor{\binits{A.}~\bsnm{Ng}},
\oauthor{\binits{D.}~\bsnm{Kleyko}} and
\oauthor{\binits{B.A.}~\bsnm{Olshausen}},
Compositional factorization of visual scenes with convolutional sparse coding
  and resonator networks,
\textit{arXiv preprint arXiv:2404.19126}
(2024).
\end{botherref}
\endbibitem

\bibitem{ShuffleNetV2_ECCV2018}
\begin{bchapter}
\bauthor{\binits{N.}~\bsnm{Ma}},
\bauthor{\binits{X.}~\bsnm{Zhang}},
\bauthor{\binits{H.-T.}~\bsnm{Zheng}} and
\bauthor{\binits{J.}~\bsnm{Sun}},
\bctitle{ShuffleNet V2: Practical Guidelines for Efficient CNN Architecture
  Design},
in: \bbtitle{Proceedings of the European Conference on Computer Vision (ECCV)},
\byear{2018}.
\end{bchapter}
\endbibitem

\bibitem{MobileNetV2_CVPR2018}
\begin{bchapter}
\bauthor{\binits{M.}~\bsnm{Sandler}},
\bauthor{\binits{A.}~\bsnm{Howard}},
\bauthor{\binits{M.}~\bsnm{Zhu}},
\bauthor{\binits{A.}~\bsnm{Zhmoginov}} and
\bauthor{\binits{L.-C.}~\bsnm{Chen}},
\bctitle{MobileNetV2: Inverted Residuals and Linear Bottlenecks},
in: \bbtitle{Proceedings of the IEEE Conference on Computer Vision and Pattern
  Recognition (CVPR)},
\byear{2018}.
\end{bchapter}
\endbibitem

\bibitem{Hersche_2022_CVPR}
\begin{bchapter}
\bauthor{\binits{M.}~\bsnm{Hersche}},
\bauthor{\binits{G.}~\bsnm{Karunaratne}},
\bauthor{\binits{G.}~\bsnm{Cherubini}},
\bauthor{\binits{L.}~\bsnm{Benini}},
\bauthor{\binits{A.}~\bsnm{Sebastian}} and
\bauthor{\binits{A.}~\bsnm{Rahimi}},
\bctitle{Constrained Few-Shot Class-Incremental Learning},
in: \bbtitle{Proceedings of the IEEE/CVF Conference on Computer Vision and
  Pattern Recognition (CVPR)},
\byear{2022},
pp.~\bfpage{9057}--\blpage{9067}.
\end{bchapter}
\endbibitem

\bibitem{hoffer2018fix}
\begin{bchapter}
\bauthor{\binits{E.}~\bsnm{Hoffer}},
\bauthor{\binits{I.}~\bsnm{Hubara}} and
\bauthor{\binits{D.}~\bsnm{Soudry}},
\bctitle{Fix your classifier: the marginal value of training the last weight
  layer},
in: \bbtitle{International Conference on Learning Representations (ICLR)},
\byear{2018}.
\end{bchapter}
\endbibitem

\bibitem{NeuralCollaps_NIPS2021}
\begin{bchapter}
\bauthor{\binits{Z.}~\bsnm{Zhu}},
\bauthor{\binits{T.}~\bsnm{Ding}},
\bauthor{\binits{J.}~\bsnm{Zhou}},
\bauthor{\binits{X.}~\bsnm{Li}},
\bauthor{\binits{C.}~\bsnm{You}},
\bauthor{\binits{J.}~\bsnm{Sulam}} and
\bauthor{\binits{Q.}~\bsnm{Qu}},
\bctitle{A Geometric Analysis of Neural Collapse with Unconstrained Features},
in: \bbtitle{Advances in Neural Information Processing Systems},
\byear{2021}.
\end{bchapter}
\endbibitem

\bibitem{mettes2019hyperspherical}
\begin{botherref}
\oauthor{\binits{P.}~\bsnm{Mettes}},
\oauthor{\binits{E.}~\bsnm{van~der Pol}} and
\oauthor{\binits{C.}~\bsnm{Snoek}},
Hyperspherical prototype networks,
\textit{Advances in Neural Information Processing Systems (NeurIPS)}
\textbf{32}
(2019).
\end{botherref}
\endbibitem

\bibitem{RandomClassVec2021}
\begin{botherref}
\oauthor{\binits{G.}~\bsnm{Shalev}},
\oauthor{\binits{G.L.}~\bsnm{Shalev}} and
\oauthor{\binits{Y.}~\bsnm{Keshet}},
Redesigning the Classification Layer by Randomizing the Class Representation
  Vectors,
\textit{arXiv preprint arXiv:2011.08704}
(2021).
\end{botherref}
\endbibitem

\bibitem{snoek2012practical}
\begin{botherref}
\oauthor{\binits{J.}~\bsnm{Snoek}},
\oauthor{\binits{H.}~\bsnm{Larochelle}} and
\oauthor{\binits{R.P.}~\bsnm{Adams}},
Practical bayesian optimization of machine learning algorithms,
\textit{Advances in Neural Information Processing systems (NeurIPS)}
\textbf{25}
(2012).
\end{botherref}
\endbibitem

\bibitem{scott2021mises}
\begin{bchapter}
\bauthor{\binits{T.R.}~\bsnm{Scott}},
\bauthor{\binits{A.C.}~\bsnm{Gallagher}} and
\bauthor{\binits{M.C.}~\bsnm{Mozer}},
\bctitle{von Mises-Fisher Loss: An Exploration of Embedding Geometries for
  Supervised Learning},
in: \bbtitle{Proceedings of the IEEE/CVF International Conference on Computer
  Vision (CVPR)},
\byear{2021}.
\end{bchapter}
\endbibitem

\bibitem{krizhevsky2009learning}
\begin{botherref}
\oauthor{\binits{A.}~\bsnm{Krizhevsky}},
Learning multiple layers of features from tiny images,
\textit{University of Toronto}
(2009).
\end{botherref}
\endbibitem

\bibitem{he2016deep}
\begin{bchapter}
\bauthor{\binits{K.}~\bsnm{He}},
\bauthor{\binits{X.}~\bsnm{Zhang}},
\bauthor{\binits{S.}~\bsnm{Ren}} and
\bauthor{\binits{J.}~\bsnm{Sun}},
\bctitle{Deep residual learning for image recognition},
in: \bbtitle{Proceedings of the IEEE Conference on Computer Vision and Pattern
  Recognition (CVPR)},
\byear{2016},
pp.~\bfpage{770}--\blpage{778}.
\end{bchapter}
\endbibitem

\bibitem{arcface_19}
\begin{bchapter}
\bauthor{\binits{J.}~\bsnm{Deng}},
\bauthor{\binits{J.}~\bsnm{Guo}},
\bauthor{\binits{N.}~\bsnm{Xue}} and
\bauthor{\binits{S.}~\bsnm{Zafeiriou}},
\bctitle{ArcFace: Additive angular margin loss for deep face recognition},
in: \bbtitle{Proceedings of the IEEE Conference on Computer Vision and Pattern
  Recognition (CVPR)},
\byear{2019},
pp.~\bfpage{4690}--\blpage{4699}.
\end{bchapter}
\endbibitem

\bibitem{zhu2017prune}
\begin{botherref}
\oauthor{\binits{M.}~\bsnm{Zhu}} and
\oauthor{\binits{S.}~\bsnm{Gupta}},
To prune, or not to prune: exploring the efficacy of pruning for model
  compression,
\textit{arXiv preprint arXiv:1710.01878}
(2017).
\end{botherref}
\endbibitem

\bibitem{saadabadi2024hyperspherical}
\begin{botherref}
\oauthor{\binits{M.S.E.}~\bsnm{Saadabadi}},
\oauthor{\binits{A.}~\bsnm{Dabouei}},
\oauthor{\binits{S.R.}~\bsnm{Malakshan}} and
\oauthor{\binits{N.M.}~\bsnm{Nasrabad}},
Hyperspherical Classification with Dynamic Label-to-Prototype Assignment,
\textit{arXiv preprint arXiv:2403.16937}
(2024).
\end{botherref}
\endbibitem

\bibitem{wah2011caltech}
\begin{botherref}
\oauthor{\binits{C.}~\bsnm{Wah}},
\oauthor{\binits{S.}~\bsnm{Branson}},
\oauthor{\binits{P.}~\bsnm{Welinder}},
\oauthor{\binits{P.}~\bsnm{Perona}} and
\oauthor{\binits{S.}~\bsnm{Belongie}},
The {C}altech-{UCSD} {B}irds-200-2011 {D}ataset
(2011).
\end{botherref}
\endbibitem

\bibitem{ruffino2023zeroshot}
\begin{bchapter}
\bauthor{\binits{S.}~\bsnm{Ruffino}},
\bauthor{\binits{G.}~\bsnm{Karunaratne}},
\bauthor{\binits{M.}~\bsnm{Hersche}},
\bauthor{\binits{L.}~\bsnm{Benini}},
\bauthor{\binits{A.}~\bsnm{Sebastian}} and
\bauthor{\binits{A.}~\bsnm{Rahimi}},
\bctitle{Zero-shot Classification using Hyperdimensional Computing},
in: \bbtitle{2024 Design, Automation and Test in Europe Conference and
  Exhibition (DATE)},
\bpublisher{IEEE},
\byear{2024}.
\end{bchapter}
\endbibitem

\bibitem{hersche2023superposdec}
\begin{bchapter}
\bauthor{\binits{M.}~\bsnm{Hersche}},
\bauthor{\binits{Z.}~\bsnm{Opala}},
\bauthor{\binits{G.}~\bsnm{Karunaratne}},
\bauthor{\binits{A.}~\bsnm{Sebastian}} and
\bauthor{\binits{A.}~\bsnm{Rahimi}},
\bctitle{Decoding Superpositions of Bound Symbols Represented by Distributed
  Representations},
in: \bbtitle{Proceedings of the 17th International Workshop on Neural-Symbolic
  Learning and Reasoning (NeSy)},
\byear{2023}.
\end{bchapter}
\endbibitem

\bibitem{vaswani2017attention}
\begin{botherref}
\oauthor{\binits{A.}~\bsnm{Vaswani}},
\oauthor{\binits{N.}~\bsnm{Shazeer}},
\oauthor{\binits{N.}~\bsnm{Parmar}},
\oauthor{\binits{J.}~\bsnm{Uszkoreit}},
\oauthor{\binits{L.}~\bsnm{Jones}},
\oauthor{\binits{A.N.}~\bsnm{Gomez}},
\oauthor{\binits{{\L}.}~\bsnm{Kaiser}} and
\oauthor{\binits{I.}~\bsnm{Polosukhin}},
Attention is all you need,
\textit{Advances in Neural Information Processing Systems (NeurIPS)}
\textbf{30}
(2017).
\end{botherref}
\endbibitem

\bibitem{desai2022random}
\begin{barticle}
\bauthor{\binits{A.}~\bsnm{Desai}},
\bauthor{\binits{L.}~\bsnm{Chou}} and
\bauthor{\binits{A.}~\bsnm{Shrivastava}},
\batitle{Random Offset Block Embedding ({ROBE}) for compressed embedding tables
  in deep learning recommendation systems},
\bjtitle{Proceedings of Machine Learning and Systems}
\bvolume{4}
(\byear{2022}),
\bfpage{762}--\blpage{778}.
\end{barticle}
\endbibitem

\bibitem{ramsauer2021hopfield}
\begin{bchapter}
\bauthor{\binits{H.}~\bsnm{Ramsauer}},
\bauthor{\binits{B.}~\bsnm{Sch{\"a}fl}},
\bauthor{\binits{J.}~\bsnm{Lehner}},
\bauthor{\binits{P.}~\bsnm{Seidl}},
\bauthor{\binits{M.}~\bsnm{Widrich}},
\bauthor{\binits{L.}~\bsnm{Gruber}},
\bauthor{\binits{M.}~\bsnm{Holzleitner}},
\bauthor{\binits{T.}~\bsnm{Adler}},
\bauthor{\binits{D.}~\bsnm{Kreil}},
\bauthor{\binits{M.K.}~\bsnm{Kopp}},
\bauthor{\binits{G.}~\bsnm{Klambauer}},
\bauthor{\binits{J.}~\bsnm{Brandstetter}} and
\bauthor{\binits{S.}~\bsnm{Hochreiter}},
\bctitle{Hopfield Networks is All You Need},
in: \bbtitle{International Conference on Learning Representations (ICLR)},
\byear{2021}.
\end{bchapter}
\endbibitem

\bibitem{loshchilov2016sgdr}
\begin{botherref}
\oauthor{\binits{I.}~\bsnm{Loshchilov}} and
\oauthor{\binits{F.}~\bsnm{Hutter}},
SGDR: Stochastic gradient descent with warm restarts,
\textit{arXiv preprint arXiv:1608.03983}
(2016).
\end{botherref}
\endbibitem

\bibitem{RPM_ImageNet_pretraining_2020}
\begin{botherref}
\oauthor{\binits{T.}~\bsnm{Zhuo}} and
\oauthor{\binits{M.}~\bsnm{Kankanhalli}},
Solving Raven's Progressive Matrices with Neural Networks,
\textit{arXiv preprint arXiv:2002.01646}
(2020).
\end{botherref}
\endbibitem

\bibitem{Zhang_2019_adacos}
\begin{bchapter}
\bauthor{\binits{X.}~\bsnm{Zhang}},
\bauthor{\binits{R.}~\bsnm{Zhao}},
\bauthor{\binits{Y.}~\bsnm{Qiao}},
\bauthor{\binits{X.}~\bsnm{Wang}} and
\bauthor{\binits{H.}~\bsnm{Li}},
\bctitle{AdaCos: Adaptively Scaling Cosine Logits for Effectively Learning Deep
  Face Representations},
in: \bbtitle{Proceedings of the IEEE/CVF Conference on Computer Vision and
  Pattern Recognition (CVPR)},
\byear{2019}.
\end{bchapter}
\endbibitem

\bibitem{schwarz2021powerpropagation}
\begin{barticle}
\bauthor{\binits{J.}~\bsnm{Schwarz}},
\bauthor{\binits{S.}~\bsnm{Jayakumar}},
\bauthor{\binits{R.}~\bsnm{Pascanu}},
\bauthor{\binits{P.E.}~\bsnm{Latham}} and
\bauthor{\binits{Y.}~\bsnm{Teh}},
\batitle{Powerpropagation: A sparsity inducing weight reparameterisation},
\bjtitle{Advances in Neural Information Processing Systems (NeurIPS)}
\bvolume{34}
(\byear{2021}),
\bfpage{28889}--\blpage{28903}.
\end{barticle}
\endbibitem

\end{thebibliography}

\clearpage
\appendix 
\setcounter{figure}{0}
\renewcommand{\thefigure}{A\arabic{figure}}
\setcounter{table}{0}
\renewcommand{\thetable}{A\arabic{table}}

\section{Effective replacement of large FCLs}

This appendix provides more details on the effective replacement of large FCLs using the proposed BCF. 
\subsection{Overall CNN training setup}\label{appx:cnn-training}
We train all CNN architectures with SGD with architecture-specific hyperparameters, summarized in Table~\ref{tab:SGDtraining}.
The training setups for the different networks mainly differ in the chosen learning rate schedule. 
ShuffleNetV2 was trained for $400$ epochs using a learning rate initially set to 0.5 and linearly decreased towards $0$ at every epoch.
MobileNetV2 was trained with a cosine learning rate schedule~\cite{loshchilov2016sgdr}, where the learning rate is decreased based on a cosine function (single cycle) from $0.04$ to $0$ within 150 epochs. 
An additional warmup period of 5 epochs was used. 
ResNet-18 and ResNet-50 were trained with an exponential decaying learning rate schedule. 
For training the ResNet-18 on the RAVEN dataset, we started with a pre-trained model from the ImageNet-1k dataset~\cite{RPM_ImageNet_pretraining_2020}.

\subsection{Loss functions for the bipolar dense resonator network integration} \label{appx:bipolar}
We evaluated different loss functions for replacing FCL with the bipolar dense resonator networks. 
A standard cross-entropy loss operating on the cosine similarities between the query vector and the fixed bipolar vectors, scaled with a trainable inverse softmax temperature $s$~\cite{hoffer2018fix}, yielded good brute-force accuracies but notable drops when integrating the resonator networks. 
For example, when replacing the final FCL of ShuffleNetV2 on ImageNet-1K, we achieved a brute-force accuracy of $66.14\%$ but a much lower resonator network accuracy ($44.88\%$). 
Alternative adaptive cosine-based loss functions, such as AdaCos~\cite{Zhang_2019_adacos}, were ineffective too. 

Instead, we found that the fixed but configurable arcface~\cite{arcface_19} loss function is the most suitable for the resonator network integration. 
Arcface computes the angle of the target logit, adds an additive angular margin ($m$) to the target angle, and gets the target logit back again by the cosine function. 
Finally, the logits are rescaled by a fixed scaling factor ($s$) before applying the cross-entropy loss. 
To find the optimal hyperparameters ($s,m$), we conducted a grid search across a wide range of configurations ($s\in \{1,10,30,50,70\},m\in \{0.0,0.05, 0.1, 0.15 \}$). 
On ImgeNet-1K, the search yielded the parameters ($70, 0.1$) for ShuffleNetV2, ($50, 0.1$) for MobileNetV2, ($70, 0.1$) for ResNet-18, and ($70, 0.1$) for ResNet-50.
The optimal parameters for the remaining datasets with ResNet-18 were ($30, 0.1$) on CIFAR-100 and ($10, 0.1$) on RAVEN. 
The large margin separation notably increased the resonator network-based accuracy, e.g., by $9.66\%$ for ShuffleNetV2 on ImageNet-1K.

\subsection{Projection dimension $D_p$}\label{appx:d}
The main results of \name using the projection layer are reported for \gls{code} vectors with a dimension of $D_p=512$. 
Here, we show that the dimension can be flexibly varied, providing a trade-off between parameter (and therefore computation) saving and accuracy. 
We varied the dimension $D_p$ from 128 (high reduction) to 1000 (no reduction since $D_p=D_o$) for MobileNetV2 on ImageNet-1K. 
The hyperparameters of \name were kept the same for all $D_p$.
Table~\ref{tab:D} shows the results. 
With an extremely low $D_p=128$, \name shows $2.86\%$ accuracy drop compared to the baseline with trainable FCL while saving $32.8\%$ of the parameters.
With a larger $D_p \geq 512$, it yields iso-accuracy with the baseline while saving up to $18.4\%$ of the parameters. 
With $D_p=768$, \name eventually surpasses the accuracy of the baseline while saving $8.7\%$ of the parameters. 

\begin{table}[t]
\caption{Hyperparamters for CNN training.}
\label{tab:SGDtraining}
\centering
% \resizebox{\linewidth}{!}{
\begin{threeparttable}
\begin{tabular}{lcrcccccc}
\toprule
                     &                            &                               &                    &                                                        & \multicolumn{4}{c}{Learning rate schedule}                                                                                                                         \\
                     \cmidrule(r){6-9}
         \begin{tabular}[c]{@{}l@{}}Dataset/\\ architecture\end{tabular}            & \multicolumn{1}{c}{Ep.} & \multicolumn{1}{c}{Bs.} & \multicolumn{1}{c}{Mmt.} & Wd. & \multicolumn{1}{c}{\begin{tabular}[c]{@{}c@{}}Init. \\ value\end{tabular}} & Type  & \begin{tabular}[c]{@{}c@{}}Decay\\ rate\end{tabular} & \begin{tabular}[c]{@{}c@{}}Step\\ size\end{tabular} \\
                     \cmidrule(r){1-1}\cmidrule(r){2-2}\cmidrule(r){3-3}\cmidrule(r){4-4}\cmidrule(r){5-5}\cmidrule(r){6-6}\cmidrule(r){7-9}
\textbf{ImageNet-1K} &                            &                               &                                                                            &                                                                           &           &                                &                                \\
ShuffleNetV2         & 400                        & 2048                          & 4e-5  & 0.9                                                                   & 0.5                                                                       & lin    & \multicolumn{1}{c}{--}         & \multicolumn{1}{c}{--}         \\
MobileNetV2          & 150                        & 256                           & 4e-5   & 0.9                                                                     & 0.04                                                                      & cos    & \multicolumn{1}{c}{--}         & \multicolumn{1}{c}{--}         \\
ResNet-18            & 100                        & 256                           & 1e-4   & 0.9                                                                     & 0.1                                                                       & exp & 0.1                            & 30              \\
ResNet-50            & 100                        & 256                           & 1e-4   & 0.9                                                                     & 0.1                                                                       & exp & 0.1                            & 30              \\
\cmidrule(r){1-1}\cmidrule(r){2-2}\cmidrule(r){3-3}\cmidrule(r){4-4}\cmidrule(r){5-5}\cmidrule(r){6-6}\cmidrule(r){7-9}
\textbf{CIFAR-100}   &                            &                               &                                                                            &                                                                           &           &                                &                                \\
ResNet-18            & 200                        & 128                           & 5e-4    & 0.9                                                                    & 0.1                                                                       & exp & 0.2                            & 60             \\
\cmidrule(r){1-1}\cmidrule(r){2-2}\cmidrule(r){3-3}\cmidrule(r){4-4}\cmidrule(r){5-5}\cmidrule(r){6-6}\cmidrule(r){7-9}
\textbf{RAVEN}       &                            &                               &                                                                            &                                                                           &           &                                &                                \\
ResNet-18            & 100                        & 256                           & 1e-4   & 0.9                                                                     & 0.1                                                                       & exp & 0.1                            & 50 \\
\bottomrule
\end{tabular}
\begin{tablenotes}\footnotesize
\item[] \normalfont{Ep.= Epochs; Bs.= Batch size; Mmt.= Momentum; Wd.= Weight decay.}
\end{tablenotes}
\end{threeparttable}
% }
\end{table}
\begin{table}[t]
\centering
\caption{Classification accuracy (\%) on ImageNet-1K when replacing the FCL in MobileNetV2 with  \name using a projection layer with variable $D_p$. The remaining configurations ($B=4$, $F=2$) are kept constant.}
\label{tab:D}
\begin{tabular}{lrllr}
\toprule
 & $D_p$ & \multicolumn{1}{c}{BF} & Fac. & \begin{tabular}[c]{@{}l@{}}Param.\\ saving$\uparrow$\end{tabular} \\
 \cmidrule(r){1-1}\cmidrule(r){2-2}\cmidrule(r){3-3} \cmidrule(r){4-5}
Baseline & --  &      $71.57^{\pm0.13}$ &  \multicolumn{1}{c}{--}    & $0.0\%$ \\
\cmidrule(r){1-1}\cmidrule(r){2-2}\cmidrule(r){3-3} \cmidrule(r){4-5}
\multirow{5}{*}{\glspl{code}} & $128$  &  $70.55^{\pm0.10}$  & $68.71^{\pm0.12}$      &   $32.8\%$  \\
 & $256$  &  $71.15^{\pm0.07}$  & $70.64^{\pm0.13}$     &   $28.0\%$   \\
 & $512$  &  $71.69^{\pm0.11}$  & $71.41^{\pm0.10}$     &   $18.4\%$   \\
 & $768$  &  $71.80^{\pm0.09}$  & $71.64^{\pm0.11}$     &   $8.7\%$    \\
 & $1000$ &  $71.86^{\pm0.16}$  & $71.56^{\pm0.18}$     &   $0.0\%$    \\
\bottomrule
\end{tabular}
\end{table}

\subsection{Maximum number of iterations $N$}\label{appx:n}
On ImageNet-1K, the maximum number of iterations of \name with 2 codebooks is set to $N=\lfloor C/(M_1+M_2) \rfloor=15$. 
However, motivated by the extremely fast convergence on the synthetic product vectors (see Fig.~3 in main text), we show that the maximum number of iterations can be further limited to only $3$ with negligible accuracy loss while reducing the computational cost.
Table~\ref{tab:N} compares the performance on ImageNet-1K between $N=15$ and $N=3$. 
Across all architectures, our \name matches the brute-force accuracy within $0.44\%$ with $N=15$ iterations, and within $0.54\%$ with $N=3$ iterations at maximum.
In both cases, the average number of iterations is lower than the maximum $N$; thus, the factorizer converges on average before the maximum $N$ is reached.

\begin{table*}[t]
\centering
\caption{\name-based replacement approach without and with projection ($D_p=512$) when allowing \name a maximum of $N=15$ (standard) iterations, or $N=3$. 
}
\label{tab:N}
\begin{tabular}{lllllll}
\toprule
& & \multicolumn{5}{c}{GSBCs (B=4)} \\
\cmidrule(r){3-7}
 &  &  & \multicolumn{2}{c}{\name ($N=15$)} &  \multicolumn{2}{c}{\name ($N=3$)} \\
 \cmidrule(r){4-5}\cmidrule(r){6-7}
 & \multicolumn{1}{c}{\begin{tabular}[c]{@{}c@{}}Baseline\\ acc.\end{tabular}} & \multicolumn{1}{c}{\begin{tabular}[c]{@{}c@{}}BF\\ acc.\end{tabular}} & \multicolumn{1}{c}{Acc.}              & \multicolumn{1}{c}{\begin{tabular}[c]{@{}c@{}}Avg. \\ iter.\end{tabular}} & \multicolumn{1}{c}{Acc.}             & \multicolumn{1}{c}{\begin{tabular}[c]{@{}c@{}}Avg. \\ iter.\end{tabular}} \\
  \cmidrule(r){1-1}\cmidrule(r){2-2}\cmidrule(r){3-3} \cmidrule(r){4-5}\cmidrule(r){6-7}
\textbf{No projection}                                                          \\
ShuffleNetV2           & $69.22^{\pm0.20}$ &  $65.09^{\pm0.10}$ & $64.76^{\pm0.13}$   &   $7.3$ & $64.68^{\pm0.15}$  & $2.4$                                                                           \\
MobileNetV2            & $71.57^{\pm0.13}$ &   $70.00^{\pm0.07}$ & $69.76^{\pm0.13}$ &  $6.2$ &  $69.72^{\pm0.13}$ & $2.3$                                                                           \\
ResNet-18              & $70.39^{\pm0.11}$ & $68.44^{\pm0.08}$ & $68.00^{\pm0.07}$ & $6.7$ & $67.90^{\pm0.07}$ &   $2.4$                                                                        \\
ResNet-50              & $76.21^{\pm0.28}$ & $76.34^{\pm0.04}$ & $76.25^{\pm0.07}$  & $4.8$ & $76.23^{\pm0.07}$  &  $2.2$                                                                          \\
 \cmidrule(r){1-1}\cmidrule(r){2-2}\cmidrule(r){3-3} \cmidrule(r){4-5}\cmidrule(r){6-7}
 \textbf{Projection}\\
ShuffleNetV2           &  $69.22^{\pm0.20}$ &  $68.67^{\pm0.11}$ &  $68.41^{\pm0.14}$ &  $6.0$ &  $68.33^{\pm0.13}$ & $2.4$ \\
MobileNetV2            &  $71.57^{\pm0.13}$ &  $71.69^{\pm0.11}$ &  $71.49^{\pm0.10}$ & $5.5$ & $71.44^{\pm0.11}$ & $2.3$                                                                           \\
ResNet-18              &  $70.39^{\pm0.11}$ &  $69.57^{\pm0.17}$ & $69.19^{\pm0.13}$ & $6.2$ &  $69.10^{\pm0.14}$ & $2.4$                                                                      \\
ResNet-50              & $76.21^{\pm0.28}$ &   $76.72^{\pm0.06}$ & $76.56^{\pm0.08}$ & $4.6$ & $76.54^{\pm0.06}$ & $2.2$  \\
\bottomrule
\end{tabular}
\end{table*}

\subsection{Number of factors $F$}\label{appx:f}
So far, we have evaluated \name with two factors, each having codebooks of size $M_1=M_2=32$ on the ImageNet-1K dataset. 
We demonstrate its capability with $F=3$ codebooks of size $M_f=10$ each. 
Consequently, the maximum number of iterations becomes $N=\lfloor 1000/30\rfloor=33$. 
Table~\ref{tab:F} compares the classification accuracy of the factorizer for $F=2$ with $F=3$. 
The factorizer with the higher number of factors achieves similar accuracy while requiring higher iterations (11 vs. 6) compared to $F=2$. 
However, since each iteration requires fewer search operations for $F=3$ ($30$ vs. $64$ searches), the overall compute saving of the FCL remains similar. 
Our \name provides a time-space trade-off between the number of factors while keeping the accuracy and the computational cost constant: a low number of factors ($F=2$) requires more space to store the codebooks ($2\cdot32=64$ codevectors) but features lower average iterations (6), whereas a higher number of factors ($F=3$) requires less space ($3\cdot 10=30$ codevectors) but more iterations on average (11). 

\begin{table}[]
\caption{Classification accuracy (\%) on ImageNet-1K when replacing the FCL in MobileNetV2 with \name using a projection with $D_p=512$ and variable $F$. }
\label{tab:F}
% \resizebox{\linewidth}{!}{
\begin{tabular}{llllrr}
\toprule
                     & \multicolumn{1}{c}{F} & \multicolumn{1}{c}{BF} & \multicolumn{1}{c}{Fac.} & \multicolumn{1}{c}{\begin{tabular}[c]{@{}c@{}}Avg. \\ iter.\end{tabular}} & \begin{tabular}[c]{@{}c@{}}FCL\,comp.\\saving$\uparrow$\end{tabular} \\
\cmidrule(r){1-1} \cmidrule(r){2-2} \cmidrule(r){3-6}
Baseline             &  \multicolumn{1}{c}{--}         &   $71.57^{\pm0.13}$  &       \multicolumn{1}{c}{--}                   &     \multicolumn{1}{c}{--} & $0.0\%$                                                            \\
\cmidrule(r){1-1} \cmidrule(r){2-2} \cmidrule(r){3-6}
\multirow{2}{*}{\glspl{code}} &    $2$   &  $71.69^{\pm0.12}$ &   $71.49^{\pm0.10}$& $6$ & $33.4\%$ \\
                     &  $3$   &  $71.61^{\pm0.07}$ & $71.27^{\pm0.10}$ & $11$ & $35.6\%$ \\
\bottomrule
\end{tabular}
% }
\end{table}

\subsection{Comparison with weight pruning}\label{appx:pruning}
% Comparison to pruning strategies 
We compare our approach to weight pruning techniques, which mostly sparsify the weights in all layers, whereas we focus on the final FCL due to its dominance in compact networks. 
Such pruning can be similarly applied to earlier layers in addition to our method. 
For a one-to-one comparison, we prune the final FCL weights of a pretrained MobileNetV2 using iterative magnitude-based pruning~\cite{zhu2017prune}.
As shown in Table~\ref{tab:pruning}, the accuracy of the magnitude-based pruning method quickly degrades as soon as more than 95\% of the weights are set to zero. 
In fact, this susceptibility forced related works to prune the final FCL only to 90\%~\cite{schwarz2021powerpropagation}. 
In contrast, our method remains accurate in high sparsity regimes: the codebooks store only $B\cdot F\cdot M_f=4 \cdot 2 \cdot 32=256$ indices instead of $ D_p \cdot C$ values (i.e., 99.98\% zero elements) with an achieved an accuracy of 69.76\%.

\begin{table}[t]
\caption{Comparison of \name with pruning of final FCL weights of a pretrained MobileNetV2 on ImageNet-1K.}
\label{tab:pruning}
\centering
% \resizebox{\linewidth}{!}{
\begin{tabular}{lrl}
\toprule
  Approach       & Zero elements (\%) & Accuracy (\%) \\
\cmidrule(r){1-1}\cmidrule(r){2-2}\cmidrule(r){3-3}
Baseline &    \multicolumn{1}{r}{--} & $71.57^{\pm0.13}$   \\
\name      &  99.98 & $69.76^{\pm0.13}$ \\
\cmidrule(r){1-1}\cmidrule(r){2-2}\cmidrule(r){3-3}
\multirow{5}{*}{FCL pruning} &      80.00    & 71.13 \\
    & 90.00    & 70.44 \\
    & 95.00    & 69.48 \\
    & 99.00    & 65.82 \\
    & 99.98 & 18.38 \\
\bottomrule
\end{tabular}
% }
\end{table}

\end{document}